\documentclass{article} % For LaTeX2e
\usepackage{iclr2027_conference,times}

\usepackage{amsmath,amsfonts,bm}

\def\eqref#1{equation~\ref{#1}}
\def\1{\bm{1}}

\DeclareMathAlphabet{\mathsfit}{\encodingdefault}{\sfdefault}{m}{sl}
\SetMathAlphabet{\mathsfit}{bold}{\encodingdefault}{\sfdefault}{bx}{n}

\usepackage{hyperref}
\usepackage{url}
\usepackage{booktabs}
\usepackage{subcaption}
\usepackage{graphicx}
\usepackage[table]{xcolor}
\usepackage{amssymb}
\usepackage{amsthm}
\newtheorem{proposition}{Proposition}

\usepackage{tikz}
\usetikzlibrary{positioning,arrows.meta}

\title{PotentialEP: Backprop-Free Training of CIFAR-Scale Generative Models via Equilibrium Propagation}

\author{Alex Gower\thanks{Correspondence: apg59@cam.ac.uk, or permanently at alexanderpgower@gmail.com} \\
University of Cambridge \& Nokia Bell Labs \\
}

\iclrfinalcopy % Uncomment for camera-ready version, but NOT for submission.
\begin{document}
\hypersetup{pdftitle={PotentialEP: Backprop-Free Training of CIFAR-Scale Generative Models via Equilibrium Propagation}, pdfauthor={Alex Gower}}

\maketitle

\begin{abstract}
Generating an image with a diffusion or flow-matching model typically involves integrating a velocity
field for hundreds of sequential steps. That computation suits a GPU poorly and an
analog neuromorphic substrate well, where one fast, low-power physical
relaxation could replace the whole sequence. Two things are needed: a generative algorithm whose computation is a
relaxation, and a way to train it without backpropagation, whose global
gradient transport is hard to realise in analog. Energy Matching supplies the first,
sampling by gradient-descent relaxation on a time-independent potential --- but
obtaining the gradient that descent follows is itself a backward pass, once
per integration step. We prove that a predictive coding network with a single linear
injection term added to its energy relaxes to a state whose visible layer
carries that gradient inherently, with an error that vanishes for small
injection strengths. An Energy Matching model trained
by backpropagation can therefore be loaded into this network unchanged and sampled by
relaxation alone, matching the original model's FID to within $0.04$ at
CIFAR-10 and ImageNet-32 scale. The network can also be trained without
backpropagation: Equilibrium Propagation reproduces backpropagation's
parameter updates to a cosine similarity of $0.9998$, and training end to end
with those updates reaches an FID of $3.64$ on unconditional CIFAR-10 against
$3.54$ for backpropagation. To our knowledge this is the first CIFAR-10
generative model trained with Equilibrium Propagation, and the first at
competitive quality by any backpropagation-free rule. This quality does not
require group normalisation or attention: an architecture of only convolutions
and pointwise nonlinearities, both native to an analog crossbar, reaches an FID of $3.51$ when trained by backpropagation.
\end{abstract}

\section{Introduction}\label{sec:introduction}

Modern generative models are expensive to train~\citep{strubellEnergyPolicyConsiderations2019}
and expensive to sample~\citep{luccioniPowerHungryProcessing2024}: diffusion, flow matching and
related methods typically produce one image by integrating a velocity field
over hundreds of sequential steps, each a full pass through a large network.
Sequential work of this kind suits a GPU poorly and a physical system well. A substrate that relaxes to the minimum of an energy performs in one continuous
settling what a GPU performs as hundreds of discrete evaluations, and analog
neuromorphic devices promise large gains in speed and energy on that basis. Two things stand in the way: finding a generative algorithm whose computation
is a relaxation, and removing the backpropagation that
training relies on.

Energy Matching \citep{balcerakEnergyMatchingUnifying2025} is one such algorithm. It generates by gradient descent on
a potential $V_\theta(x)$, the scalar output of a feedforward network, whose
minima are where the data lies. The potential does not change with time, so
every step descends the same fixed landscape, which is what a physical
system does when it relaxes. But the velocity it follows is $-\nabla_x V_\theta$, and in its original
formulation obtaining that gradient is a backward pass --- once per
integration step during sampling, on top of the usual $\nabla_\theta \mathcal{L}$ backward pass per training step.

The remaining obstacle is therefore backpropagation, whose global gradient
transport is difficult to realise in analog.
Mixed-signal crossbar designs convert between analog and digital at every
layer, at a cost in energy and latency \citep{kendallTrainingEndtoEndAnalog2020}. Physics-aware training instead runs the forward pass physically but the backward pass
through a digital model of the substrate, and pays for digital computation at
every training step \citep{wrightDeepPhysicalNeural2022}. Both pay a digital
cost the physics does not.

\paragraph{The correspondence.}
We show that the physical dynamics of a predictive coding network can compute $-\nabla_x V_\theta$ inherently. Build a PCN from the same layers, one prediction error per
node, and add a single linear term of strength $\gamma$ on the scalar output
node, giving an energy $E_{\rm PCN}$. The states of this \emph{PotentialPCN}
relax to an equilibrium at which $-\nabla_x E_{\rm PCN} =
\gamma\,(-\nabla_x V_\theta) + O(\gamma^2)$, so dividing by $\gamma$ recovers
the velocity, exactly in the limit of small $\gamma$
(Section~\ref{subsec:correspondence}).

\paragraph{Generation without a backward pass.}
An Energy Matching model trained conventionally by
backpropagation can therefore be loaded into this network unchanged and sampled by
relaxation, so that a substrate hosting it needs no backward pass anywhere in the generation path. At CIFAR-10
and ImageNet-32 scale the velocities agree with autograd's to a cosine
similarity of $1$ to nine decimal places, and the FID of the generated samples matches to
within $0.04$ (Section~\ref{subsec:equivalence-at-inference}). No retraining is required, so the energy and latency of generation --- most of
a deployed model's lifetime cost --- could be cut on analog hardware without
any change to how the model is trained.

\paragraph{Training without a backward pass.}
Training can move on-chip as well. The PCN relaxes on an
explicit energy, so it can be trained by Equilibrium Propagation
\citep{scellierEquilibriumPropagationBridging2017}, which reads parameter
updates $-\nabla_\theta
\mathcal{L}$ from local measurements at equilibrium and so needs none of the global
transport that makes backpropagation awkward in analog. We use the GradEP
variant \citep{gowerFlowEqPropTrainingFlow2026}, which trains gradients of
energy landscapes --- here the velocities Energy Matching generates with. We
call the resulting procedure \emph{PotentialEP}. Its parameter updates reproduce
backpropagation's to a cosine similarity of $0.9998$ at our operating point
(Section~\ref{subsec:equivalence-at-training}), and a model trained end to end
with those updates reaches an FID of $3.64$ on unconditional CIFAR-10, against
$3.54$ for backpropagation (Section~\ref{sec:cifar}). To our knowledge this is
the first CIFAR-10 generative model trained with Equilibrium Propagation, and
the first at competitive quality by any backpropagation-free rule.

\paragraph{The architecture.}
A backpropagation-free algorithm is of little use if the architecture it trains
is itself hard to realise in analog, and two components of the one we inherit
from Energy Matching are: group normalisation, which couples every channel
through a shared statistic, and attention, whose pairwise products are not a
fixed weighted sum. We find that removing both costs nothing measurable at this
scale --- a network of only convolutions, pointwise nonlinearities and UNet
skips, every operation of the kind an analog crossbar performs natively,
reaches a CIFAR-10 FID of $3.51$ (Section~\ref{subsec:hardware}).

\section{Background}
\label{sec:background}

\subsection{Energy Matching}
\label{subsec:bg-energy-matching}

Energy Matching \citep{balcerakEnergyMatchingUnifying2025} generates images by
relaxation on a learned scalar potential. A feedforward network with a scalar
output defines
\begin{equation}\label{eq:em-potential}
    V_\theta(x) = \alpha\, F_\theta(x) ,
\end{equation}
where $F_\theta(x)$ is the network output and $\alpha$ a fixed output scale set
by the architecture, and samples are produced by integrating $\dot x =
-\nabla_x V_\theta(x)$ from noise. Unlike diffusion or flow matching, the
velocity field is the gradient of a \emph{time-independent} potential: the same
landscape is descended at every step, and the model's minima are where the data
lies.

Training has two phases. The first is flow matching
\citep{lipmanFlowMatchingGenerative2023} on exact optimal-transport couplings
\citep{tongImprovingGeneralizingFlowbased2024}: for a noise sample $x_0$ paired
with a data sample $x_1$, an interpolant $x_t = (1-t)x_0 + tx_1$ with $t \sim
\mathcal{U}[0,1)$, and a target velocity $u = x_1 - x_0$, the loss is
\begin{equation}\label{eq:fm-loss}
    \mathcal{L}(\theta) = \tfrac12 \big\|{-\nabla_x V_\theta(x_t)} - u\big\|^2 .
\end{equation}
The coupling is needed because the field $-\nabla_x V_\theta$ is
time-independent and curl-free by construction, while the marginal velocity it
is asked to match need be neither; pairing each $x_0$ with its transport-optimal $x_1$ reduces the overlap between
conditional trajectories, so that the marginal they define is close to
time-independent and curl-free. The second phase adds a
contrastive term, $\nabla_\theta V(x^+) - \nabla_\theta V(x^-)$ between data
samples and model samples produced by a short Langevin chain. We refer to these throughout as
phases 1 and 2.

Sampling integrates
\begin{equation}\label{eq:em-sampler}
    \mathrm{d}x = -\nabla_x V_\theta(x)\,\mathrm{d}\tau + \sqrt{\epsilon(\tau)}\,\mathrm{d}W
\end{equation}
to a stopping time $\tau_s$, with the noise $\epsilon$ switched off below
$\tau = 1$ and held at a fixed level beyond it: up to $\tau = 1$ it is a
noiseless ODE transporting noise to data, and beyond that a Langevin chain at
fixed temperature. Because the landscape does not change with time,
integrating past $\tau = 1$ is further equilibration rather than extrapolation
into an untrained regime, as it would be for a time-conditioned flow-matching
model, so $\tau_s$ is a free parameter, chosen per model over a ladder of
values (App.~\ref{app:experimental-implementation}).

\subsection{Predictive coding networks}
\label{subsec:bg-pcn}

A predictive coding network~\citep{raoPredictiveCodingVisual1999, whittingtonApproximationErrorBackpropagation2017} assigns a dynamical variable $h_k$ to each node of a feedforward
architecture and an energy to the mismatch between each node and the
prediction made of it by its predecessor:
\begin{equation}\label{eq:pcn-plain}
    E = \sum_{k=1}^{N} \tfrac12 \big\| f_k(h_{k-1}) - h_k \big\|^2 ,
    \qquad e_k = f_k(h_{k-1}) - h_k .
\end{equation}
Inference is not a forward pass but a relaxation: with the input $h_0 = x$
clamped, the states descend $E$ to a stationary point, and outputs are read
from the settled state. Two properties of that stationary state are what
Section~\ref{subsec:correspondence} builds on. First, with only error terms
present the minimum is $e = 0$, which by induction from $h_0 = x$ is exactly
the forward pass: left to itself the relaxation reproduces the feedforward
network it was built from. Second, stationarity $\partial E/\partial h_k = 0$
at an interior node gives $e_k = J_{k+1}^{\top} e_{k+1}$ with $J_{k+1} =
\partial f_{k+1}/\partial h_k$, so any error introduced at the top propagates
downwards through transposed Jacobians. 

Each term of \eqref{eq:pcn-plain} involves only a node and its predecessor, so
both the relaxation and the parameter gradients $\partial E/\partial\theta_k =
e_k^{\top}\,\partial f_k/\partial\theta_k$ are local: every quantity a weight
needs is present at its two ends. This is what makes predictive coding a
candidate for analog implementation, where the relaxation is performed by the
physics rather than simulated.

\subsection{Equilibrium Propagation}
\label{subsec:bg-ep}

Equilibrium Propagation \citep{scellierEquilibriumPropagationBridging2017,
laborieuxScalingEquilibriumPropagation2020} is an alternative learning algorithm to
backpropagation for systems whose dynamics settle to stationary points, from
which their outputs are read. It uses only local measurements at equilibrium,
making it a natural candidate for neuromorphic hardware.

Consider a system with state $s$, parameters $\theta$ and energy
$E(s;\theta)$, relaxing to an equilibrium $s^*$ whose position we want to train
so that the outputs reduce a cost $C(s)$. Adding $\beta C(s)$ to the energy
displaces that equilibrium towards lower cost, and the parameter gradient can
be read from where and how far it moves. Each training sample therefore takes three
phases: a \emph{free phase} relaxing to $s^*$, then two \emph{nudged phases}
relaxing $E \pm \beta C$ to $s^{*\pm\beta}$, at lower and higher cost
respectively. The update is
\begin{equation}\label{eq:ep-theorem}
    \Delta\theta \;\propto\; -\frac{1}{2\beta}
    \Big[\frac{\partial E}{\partial\theta}\Big|_{s^{*+\beta}}
       - \frac{\partial E}{\partial\theta}\Big|_{s^{*-\beta}}\Big]
    \;\xrightarrow[\beta \to 0]{}\; -\nabla_\theta\, C(s^*) ,
\end{equation}
which lowers the energy at $s^{*+\beta}$ and raises it at $s^{*-\beta}$, so
that the free equilibrium $s^*$ itself shifts towards lower cost. For small
enough $\beta$ this is exactly gradient descent on $C(s^*)$.

Each $\partial E/\partial\theta$ is a local quantity --- on a PCN,
$e_k^{\top}\partial f_k/\partial\theta_k$ --- so the whole procedure runs
without global backward passes. EP has been scaled to CIFAR-10 and ImageNet
classification, including on PCN
energies~\citep{kerjanTrainingPredictiveCoding2026}, but its generative
applications remain at MNIST scale and below
\citep{meerschTrainingHopfieldVariational2023, gowerFlowEqPropTrainingFlow2026}.

The cost $C$ must be a function of $s^*$ itself, the position the state
settles at, since that is what the nudge displaces. A loss involving an energy
\emph{gradient} at that position --- such as Energy Matching's
\eqref{eq:fm-loss} involving $\nabla_x V$ --- is therefore out of reach.
GradEP \citep{gowerFlowEqPropTrainingFlow2026} resolves this by running EP on
a modified energy whose equilibrium position encodes the gradient of the
original energy, so that moving the former's equilibrium trains the latter's
gradient. Section~\ref{subsec:training} applies this to the PCN of
Section~\ref{subsec:correspondence}, and
Section~\ref{subsec:linearised-nudge} improves the nudge it uses.

\section{Method}
\label{sec:method}

\subsection{A PCN whose relaxation computes \texorpdfstring{$-\nabla_x V$}{-grad V}}
\label{subsec:correspondence}

Energy Matching defines its potential as the scalar output of a feedforward
network, scaled by a fixed constant $\alpha$: $V_\theta(x) = \alpha\,
F_\theta(x)$, with $F_\theta = f_N \circ \cdots \circ f_1$ the network and
$F_\theta(x)$ its scalar output. Generation integrates $\dot x = -\nabla_x V$,
and every evaluation of that gradient is a backward pass through $F_\theta$. We
now build, from the same layers $f_k$, a predictive coding network that
produces the same gradient as a property of its relaxed state, reached by
dynamics that are local at every node (Figure~\ref{fig:schematic}).

Take the standard PCN energy for this network --- one prediction error per node
--- and add a single linear \emph{injection} term on the output node:
\begin{equation}\label{eq:pcn-energy}
    E_{\rm PCN}(x, h) \;=\; \sum_{k=1}^{N} \tfrac12 \big\| f_k(h_{k-1}) - h_k \big\|^2
    \;+\; \gamma\,\alpha\, h_N ,
    \qquad h_0 = x,\;\; h_N = o \in \mathbb{R},
\end{equation}
where $h = (h_1, \dots, h_N)$ are the node states, $h_N = o$ is the scalar
output node ($\alpha$ is kept separate from $\gamma$ so that $\gamma > 0$ is
expressed on the potential's scale), and $e_k = f_k(h_{k-1}) - h_k$ are the
errors. We call this construction a
\emph{PotentialPCN}. The hidden states relax at fixed $x$
to a stationary point $h^*(x)$ of $E_{\rm PCN}$, and we read a velocity from
the visible layer:
\begin{equation}\label{eq:vpcn}
    v_{\rm PCN}(x) \;=\; -\,\gamma^{-1}\, \nabla_x E_{\rm PCN}\big(x, h^*(x)\big) .
\end{equation}

\begin{proposition}\label{prop:correspondence}
For smooth $f_k$ and $\gamma \to 0$,
\begin{equation}
    v_{\rm PCN}(x) = -\nabla_x V_\theta(x) + O(\gamma),
    \qquad
    \gamma^{-1} E_{\rm PCN}\big(x, h^*(x)\big) = V_\theta(x) + O(\gamma).
\end{equation}
\end{proposition}

\begin{figure}[t]
\centering
\footnotesize
\begin{tikzpicture}[
  node/.style={draw, rounded corners=2pt, minimum width=7mm, minimum height=5mm, inner sep=1pt},
  state/.style={node, fill=blue!7},
  ebox/.style={draw, rounded corners=2pt, fill=gray!10, inner sep=3pt, align=center, font=\footnotesize},
  err/.style={draw, circle, minimum size=5mm, inner sep=0pt, fill=gray!10},
  inj/.style={draw, minimum size=4.2mm, inner sep=1.5pt, fill=gray!10, font=\footnotesize},
  fwd/.style={-{Latex[length=1.5mm]}, thick},
  bwd/.style={-{Latex[length=1.5mm]}, thick, dashed},
  dep/.style={gray!90, thin, dashed},
  lbl/.style={font=\footnotesize, inner sep=1pt},
  dots/.style={font=\footnotesize, inner sep=3pt},
  dyn/.style={font=\footnotesize, inner sep=1pt, gray!80!black},
  dynbox/.style={dyn, draw=gray!60, dotted, rounded corners=3pt, inner sep=2.5pt, fill=gray!4},
  panel/.style={font=\normalsize, inner sep=1pt, anchor=west},
  x=1cm, y=1cm]

%%%%%%%%%%%%%%%%%%%%%%%% (a) Energy Matching %%%%%%%%%%%%%%%%%%%%%%%%
\begin{scope}
  \node[panel] at (-2.2,1.8) {\textbf{(a)} Energy Matching};
  \node[node] (x)  at (0,0)   {$x$};
  \node[node] (h1) at (1.4,0) {$h_1$};
  \node[node] (h2) at (2.8,0) {$h_2$};
  \node[dots] (dots) at (3.9,0) {$\cdots$};
  \node[node] (hN) at (5.4,0) {$h_N$};
  \node[ebox] (V) at (7.3,0) {$V = \alpha h_N$};
  \draw[fwd] (x) -- node[lbl, above=2pt] {$f_1$} (h1);
  \draw[fwd] (h1) -- node[lbl, above=2pt] {$f_2$} (h2);
  \draw[fwd] (h2) -- (dots);
  \draw[fwd] (dots) -- node[lbl, above=2pt] {$f_N$} (hN);
  \draw[fwd] (hN) -- (V);
  \node[dynbox, above=8pt of x] {$\dot x = -\nabla_x V$};
  \draw[bwd] (V.south) -- ++(0,-1.35) -| (x.south)
        node[lbl, pos=0.25, below=2pt] {backward pass through every layer};
  \node[lbl] at (3.65,-2.4)
        {$\nabla_x V = \alpha\, J_1^{\top} J_2^{\top}\cdots J_N^{\top}$};
\end{scope}

%%%%%%%%%%%%%%%%%%%%%%%% (b) PotentialPCN %%%%%%%%%%%%%%%%%%%%%%%%
\begin{scope}[yshift=-5.2cm]
  \node[panel] at (-2.2,1.8) {\textbf{(b)} PotentialPCN};
  % free hidden states (shaded)
  \node[node]  (x)  at (0,0)   {$x$};
  \node[state] (h1) at (1.4,0) {$h_1$};
  \node[state] (h2) at (2.8,0) {$h_2$};
  \node[dots]  (dots) at (3.9,0) {$\cdots$};
  \node[state] (hN) at (5.4,0) {$h_N$};
  % error units below their states
  \node[err]  (e1) at (1.4,-1.4) {$e_1$};
  \node[err]  (e2) at (2.8,-1.4) {$e_2$};
  \node[dots] (edots) at (3.9,-1.4) {$\cdots$};
  \node[err]  (eN) at (5.4,-1.4) {$e_N$};
  % predictions go to the error units
  \draw[fwd] (x.south east)  -- node[lbl, above right] {$f_1$} (e1.north west);
  \draw[fwd] (h1.south east) -- node[lbl, above right] {$f_2$} (e2.north west);
  \draw[fwd] (h2.south east) -- (edots.north);
  \draw[fwd] (dots.south east) -- node[lbl, above right] {$f_N$} (eN.north west);
  % which compare them against the states
  \draw[-] (h1) -- node[lbl, right] {$\ominus$} (e1);
  \draw[-] (h2) -- node[lbl, right] {$\ominus$} (e2);
  \draw[-] (hN) -- node[lbl, right] {$\ominus$} (eN);
  % dynamics, both above the chain
  \node[dynbox, above=8pt of x] {$\dot x = -\gamma^{-1}\nabla_x E_{\rm PCN}(x, h^*)$};
  \node[dyn] at (4.0,0.6) {$\dot h_k = -\nabla_{h_k} E_{\rm PCN}$};
  % injection term to the right of h_N
  \node[inj] (I) at (8.0,0) {$\gamma\alpha h_N$};
  \draw[-] (hN) -- (I);
  \node[dyn, above=4pt of I] {$\approx \gamma V$ at $h^*$};
  % energy, fed by the error units and the injection
  \node[ebox] (E) at (8.0,-1.4) {$E_{\rm PCN} = \sum_k \tfrac12\|e_k\|^2 + \gamma\alpha\, h_N$};
  \draw[dep] (eN) -- (E);
  \draw[dep] (I.south) -- (E.north);
  % equilibrium recursion along the error units; labels on the dashed chain
  \draw[bwd] (eN) -- node[lbl, above=2pt] {$J_N^{\top}$} (edots);
  \draw[bwd] (edots) -- (e2);
  \draw[bwd] (e2) -- node[lbl, above=2pt] {$J_2^{\top}$} (e1);
  \draw[bwd] (e1) -| node[lbl, pos=0.25, above=2pt] {$J_1^{\top}$} (x.south);
  \node[lbl, below=3pt of eN] {$e_N^* = \gamma\alpha$};
  \node[lbl] at (3.2,-2.4)
        {$\nabla_x E_{\rm PCN}(x, h^*) = J_1^{\top} e_1^*
          \approx \gamma\alpha\, J_1^{\top} J_2^{\top}\cdots J_N^{\top}
          = \gamma\,\nabla_x V$};
\end{scope}
\end{tikzpicture}
\caption{The same network, read two ways. \textbf{(a)} Energy Matching
generates by $\dot x = -\nabla_x V$, and each evaluation of $\nabla_x V$ is a
backward pass through the network ($J_k = \partial f_k/\partial h_{k-1}$).
\textbf{(b)} In the PotentialPCN the same blocks $f_k$ make predictions, but
the hidden states (shaded) are not computed from them: instead they are free
variables that relax by descending $E_{\rm PCN}$ to stationary values $h^*$.
Each error unit $e_k = f_k(h_{k-1}) - h_k$ compares a prediction against a
state, and the energy is the sum of squared errors plus a single linear term
$\gamma\alpha h_N$, which at $h^*_N$ equals $\gamma V$ up to $O(\gamma^2)$.
The same term fixes $e_N^* = \gamma\alpha$, and the recursion
$e_k^* = J_{k+1}^{\top}e_{k+1}^*$ carries it down through the same transposed
Jacobians as (a), so the gradient of $E_{\rm PCN}$ at the visible node is
$\gamma\,\nabla_x V + O(\gamma^2)$. Dividing by $\gamma$, the visible
dynamics $\dot x = -\gamma^{-1}\nabla_x E_{\rm PCN}$ reproduce those of
(a) to $O(\gamma)$: the backward pass has been performed by the relaxation.}
\label{fig:schematic}
\end{figure}

The argument is short, and App.~\ref{app:derivation} gives it in full. Without
the injection the energy is minimised at $e = 0$, which by induction from
$h_0 = x$ is the forward pass. The injection disturbs this: stationarity in the
output node gives $e_N^* = \gamma\alpha$ exactly, and stationarity in an
interior node gives $e_k^* = J_{k+1}^{\top} e_{k+1}^*$ with $J_{k+1} =
\partial f_{k+1}/\partial h_k$, so every error is $O(\gamma)$ and each node
$h_k^*$ sits $O(\gamma)$ from the forward-pass state it would occupy at
$\gamma = 0$. The velocity is the total derivative of
$E_{\rm PCN}(x, h^*(x))$ in $x$, which has an explicit part and a part through
$h^*(x)$; the latter vanishes because $h^*$ is stationary in $h$, and iterating
the interior recursion from the top leaves $\nabla_x E_{\rm PCN}(x, h^*(x)) =
J_1^{\top} e_1^* = \gamma\alpha\, J_1^{\top} J_2^{\top} \cdots J_N^{\top}$.
Evaluating each Jacobian at the forward pass rather than at $h^*$ costs
$O(\gamma)$, and the product is then the chain rule for $\nabla_x F_\theta$, so
$\nabla_x E_{\rm PCN}(x, h^*(x)) = \gamma\alpha\,\nabla_x F_\theta(x) +
O(\gamma^2)$. For the second claim, evaluate $E_{\rm PCN}$ at $h^*$: the
squared errors are $O(\gamma^2)$, so the injection term dominates, giving
$\gamma V_\theta(x) + O(\gamma^2)$.

The PCN's errors \emph{are} the backward pass:
the linear term injects the constant $\gamma\alpha$ at the output, analogous to
the ``$\partial V/\partial V = 1$'' that autograd seeds its backward pass with,
and the relaxation propagates it down through transposed Jacobians, exactly as
backpropagation would. The velocity is then read from the first node alone,
$-\gamma^{-1} J_1^{\top} e_1^*$: where backpropagation must traverse all $N$
layers to reach $x$, here that traversal has already happened in the
relaxation, and the visible layer sees only its own local error. 

Nothing in the derivation used the values of the weights: it holds for any
$\theta$, so a network trained conventionally as $F_\theta$ can be loaded into
\eqref{eq:pcn-energy} unchanged and sampled by relaxation
(Section~\ref{subsec:equivalence-at-inference}). What the derivation does
assume is that $h$ is at equilibrium for the current $x$. In simulation we
enforce this by alternating: $x$ takes an integration step, then $h$ is relaxed
by $K_h$ sweeps of gradient descent on $E_{\rm PCN}$ at step $\mathrm{d}t$
(which in the error parameterisation of App.~\ref{app:error-param} converges in
a single sweep $K_h = 1$ at $\mathrm{d}t = 1$). A physical realisation needs no
alternation: if the hidden nodes relax on a faster timescale than $x$, $h$
tracks $h^*(x)$ adiabatically and the generation trajectory is one continuous
relaxation. The generalisation from a chain to the UNet's skip-connected graph
replaces $J_{k+1}^{\top} e_{k+1}$ by a sum over children and is given in
App.~\ref{app:dag-pcn}.

\subsection{Training the PCN by Equilibrium Propagation}\label{subsec:training}

Section~\ref{subsec:correspondence} shows that the PCN's relaxation reproduces
the generation velocity $-\nabla_x V_\theta$ for any weights $\theta$, whether
or not those weights were themselves trained on the PCN. Training them on the
PCN requires two more things: that the correspondence hold for the loss's
\emph{parameter} gradient and not only for the velocity, and a way to measure
that parameter gradient without a backward pass.

\paragraph{The PCN's loss is the potential's loss.}
In its first phase Energy Matching trains $V_\theta$ by flow matching: for an
interpolant $x_t$ between a noise sample and a data sample, with target
velocity $u$, the loss is $\mathcal{L}(\theta) = \tfrac12\|{-\nabla_x
V_\theta(x_t)} - u\|^2$, and $\Delta\theta_{\rm BP} = -\nabla_\theta
\mathcal{L}$ is its backpropagated gradient. To allow this loss (dependent on an energy \textit{gradient}) to be trained using EP, we use GradEP's spring-clamped free phase: rather than
fixing $x$ at $x_t$, we tether $x$ near $x_t$ with a spring energy term and let $x$ and $h$
relax together on
\begin{equation}\label{eq:free-energy}
    E_{\rm free}(x, h) = E_{\rm PCN}(x, h) + \tfrac{\lambda}{2}\|x - x_t\|^2 .
\end{equation}
Stationarity in $x$ gives $\nabla_x E_{\rm PCN} = -\lambda(x^* - x_t)$, so by
\eqref{eq:vpcn} the velocity we want to train is encoded in the spring's equilibrium
extension,
\begin{equation}\label{eq:velocity-readout}
    v^* \;=\; -\gamma^{-1}\nabla_x E_{\rm PCN}(x^*, h^*)
        \;=\; \tfrac{\lambda}{\gamma}\,(x^* - x_t) ,
\end{equation}
and the PCN's loss is $\mathcal{L}_{\rm PCN}(\theta) = \tfrac12\|v^* - u\|^2$.
Training is now a matter of moving where $x$ settles on this spring-clamped
energy landscape: the weights must be adjusted so that the spring's
equilibrium extension encodes a velocity closer to $u$, which is the kind of
objective EP can act on. The correspondence error varies smoothly with
$\theta$ as well as being small at each (App.~\ref{app:derivation}), so
$\nabla_\theta \mathcal{L}_{\rm PCN} = \nabla_\theta \mathcal{L} + O(\gamma)$
and descending $\mathcal{L}_{\rm PCN}$ descends the potential's own loss:
EP shapes the $E_{\rm PCN}$ landscape and $V_\theta$ is trained as a
consequence, a claim Section~\ref{subsec:equivalence-at-training} tests over
six decades of $\gamma$.

\paragraph{Two ways to compute the gradient.}
Because $h^*$ is defined implicitly by stationarity, the parameter gradient of
the flow loss on $v_{\rm PCN}(x_t)$, with $x$ held fixed at $x_t$ rather
than tethered, can be obtained by implicit differentiation, solving a linear
system in the relaxation Hessian (App.~\ref{app:error-param}). We write this update $\Delta\theta_{\rm IFT}$ and use it as the control in
Section~\ref{subsec:equivalence-at-training}: it carries none of EP's
finite-difference error, only the $O(\gamma)$ correspondence bias
(App.~\ref{app:training-fidelity}). Its matrix--vector products use autograd,
however, so a neuromorphic substrate would run Equilibrium Propagation instead.

\paragraph{The three phases of EP.}
GradEP realises \eqref{eq:ep-theorem} in three phases, using the centred
$\pm\beta$ form of \citet{laborieuxScalingEquilibriumPropagation2020}: a free
phase that relaxes \eqref{eq:free-energy} and reads $v^*$, then two nudged
phases that add $\pm\beta' C(x)$ to it and relax again from the free
equilibrium. The cost is the loss as a function of the visible state,
$C(x) = \tfrac12\|v(x) - u\|^2$ with $v(x) = \tfrac{\lambda}{\gamma}(x - x_t)$,
so that $C(x^*) = \mathcal{L}_{\rm PCN}$. Its Hessian in $x$ is
$(\lambda/\gamma)^2 I$, inherited from the velocity conversion, so we nudge
with $\beta' = \beta\,(\gamma/\lambda)^2$: the nudge then contributes $\beta I$
against the tether's $\lambda I$, making $\beta$ a stiffness directly
comparable to $\lambda$. Equation~\ref{eq:ep-theorem} then gives
\begin{equation}\label{eq:ep-update}
    \Delta\theta_{\rm EP}
    = -\,\frac{1}{2\beta'}
      \Big[\frac{\partial E_{\rm PCN}}{\partial\theta}(s^{*+\beta})
         - \frac{\partial E_{\rm PCN}}{\partial\theta}(s^{*-\beta})\Big]
    = -\nabla_\theta \mathcal{L}_{\rm PCN} + O(\beta^2),
\end{equation}
the second equality using $\nabla_\theta C(x^*) = \nabla_\theta
\mathcal{L}_{\rm PCN}$, and the size of the $O(\beta^2)$ term depending on the
choice of cost in a way Section~\ref{subsec:linearised-nudge} addresses. Only
$E_{\rm PCN}$ appears in \eqref{eq:ep-update}: the tether and the nudge are
$\theta$-free. App.~\ref{app:training-details} gives the sweep budgets, the
choice of $\lambda$, how $\mathcal{L}_{\rm PCN}$ differs from $\mathcal{L}$ at
finite $\gamma$, and the contrastive phase, which on the PCN needs no nudge at
all.

\subsubsection{The linearised nudge}\label{subsec:linearised-nudge}

GradEP's cost is the loss as a function of the visible state,
$C(x) = \tfrac12\|v(x) - u\|^2$, and being quadratic in $x$, its nudge acts as
a second spring (of strength $\beta$) against the tether (of strength
$\lambda$). Balancing the two forces gives a nudged equilibrium
$x^{*\beta} - x^* = -\beta\gamma g/\lambda(\lambda + \beta)$, with
$g = v^* - u$ (App.~\ref{app:linearised-nudge}), which is nonlinear in
$\beta$. The $\pm\beta$ displacements are therefore asymmetric about the free
equilibrium, and the centred difference \eqref{eq:ep-update} acquires an
$O((\beta/\lambda)^2)$ bias. This is standard EP's weak-nudge requirement,
with the tether's $\lambda$ in place of an ordinary network's $O(1)$
curvature. At $\lambda = 1$ it caps $\beta$ near one, which is restrictive
since the displacement the parameter update is read from scales with
$\beta\gamma$ and must stay resolvable in finite-precision arithmetic, while
$\gamma$ must stay small for the correspondence.

We therefore instead nudge with the loss's \emph{linearisation} at the free
equilibrium, as in Dual Propagation
\citep{hoierDualPropagationAccelerating2023}: freezing $g$ after the free phase
and taking $C_{\rm lin}(x) = g^{\top} v(x)$, which targets the same parameter
gradient but adds no curvature in $x$. Its gradient
$\nabla_x(\beta' C_{\rm lin}) = \beta\gamma g/\lambda$ is a constant force,
where the quadratic cost's depends on $x$ and so makes the equilibrium
nonlinear in $\beta$. The displacement
$x^{*\beta} - x^* = -\beta\gamma g/\lambda^2$ is therefore linear in
$\beta$ at any nudge strength and the $\pm\beta$ equilibria stay symmetric about the free one.
What remains is only that $\partial_\theta E_{\rm PCN}$ varies nonlinearly
along the straight path between them, so the estimate at finite $\beta$ departs from the
$\beta \to 0$ limit it targets by a relative error of order
$|x^{*\beta} - x^*|^2$ (App.~\ref{app:linearised-nudge}), and the constraint
becomes
\begin{equation}\label{eq:lin-constraint}
    \beta\,\gamma\,|g| \ll \lambda^2
    \qquad\text{in place of}\qquad
    \beta \ll \lambda ,
\end{equation}
so smaller $\gamma$ permits larger $\beta$: at the $\gamma = 10^{-4}$,
$\lambda = 1$ and $|g| \approx 0.39$ of our phase-1 training runs, the ceiling moves
from $\beta \ll 1$ to $\beta \ll 2.6\times10^4$. Section~\ref{subsec:equivalence-at-training}
measures the resulting window in $\beta\gamma$.

\section{Equivalence Experiments}

\subsection{Equivalence at Inference}\label{subsec:equivalence-at-inference}

We first compare, at identical trained weights, the velocities obtained by
automatic differentiation through the feedforward Energy Matching model,
$v_{\rm FFN}(x) = -\nabla_x V(x)$, against those produced by the relaxed PCN
built from it (App.~\ref{app:dag-pcn}), $v_{\rm PCN}(x) = -\gamma^{-1}
\nabla_x E_{\rm PCN}(x,h)|_{h^*}$. This tests the correspondence of
Section~\ref{subsec:correspondence} numerically.

\begin{figure}[tb]
\centering
\includegraphics[width=0.6\linewidth]{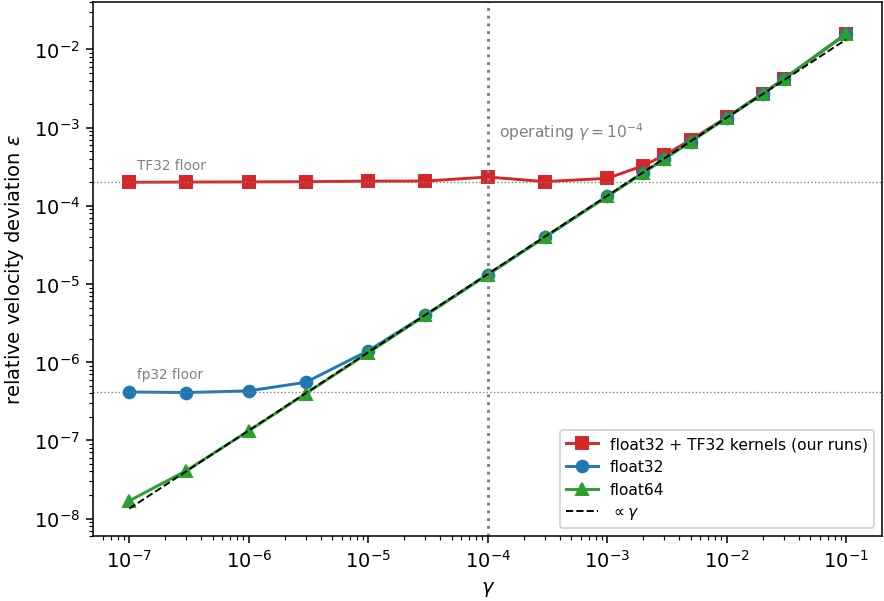}
\caption{Relative deviation
$\epsilon = \|v_{\rm PCN} - v_{\rm FFN}\| / \|v_{\rm FFN}\|$ of the relaxed
PCN velocity from the exact FFN autograd velocity, on the CIFAR-10
replication checkpoint of Table~\ref{tab:ffn2pcn-fid}, evaluated at $512$ flow-matching interpolants $x_t$ constructed from training
images exactly as during training. The deviation is $O(\gamma)$ (dashed)
until it meets the arithmetic floor of the number format; float64 meets no
floor in the measured range, and at the operating $\gamma = 10^{-4}$ (dotted)
the deviation lies below the TF32 floor of our runs.}
\label{fig:gamma-floors}
\end{figure}

\paragraph{The inference velocity correspondence is $O(\gamma)$ down to
machine precision.}
Figure~\ref{fig:gamma-floors} displays the relative velocity deviation
$\epsilon = \|v_{\rm PCN} - v_{\rm FFN}\| / \|v_{\rm FFN}\|$ as a function of $\gamma$, for our replication of the CIFAR-10 Energy
Matching model~\citep{balcerakEnergyMatchingUnifying2025}. The deviation decays proportional to $\gamma$ over six decades, from
$\gamma = 0.1$ down to $\gamma = 10^{-7}$, in both its direction and magnitude  
components (App.~\ref{app:correspondence-measurement}), as predicted by the derivation in App.~\ref{app:derivation}. At small $\gamma$ each arithmetic saturates at its own precision floor
(App.~\ref{app:correspondence-measurement}); the highest is the
$\sim 2 \times 10^{-4}$ of the TF32 tensor-core mode our runs use. Above
$\gamma \sim 10^{-2}$ the curves coincide, the finite-$\gamma$ bias having
grown past every floor. At
the $\gamma = 10^{-4}$ of our CIFAR-10 runs the float64 total relative
deviation is $1.3 \times 10^{-5}$ (a direction cosine similarity of $\cos(v_{\rm FFN}, v_{\rm PCN}) = 1$ to nine
decimal places), which lies below the TF32 floor. The correspondence error for our runs is therefore dominated by standard GPU
arithmetic precision rather than by the finite-$\gamma$ bias of the
relaxation, so smaller $\gamma$ buys no accuracy in this arithmetic. Repeating these measurements on the authors' released ImageNet-32 checkpoints
gives the same behaviour (App.~\ref{app:imagenet-transfer}), so the
correspondence is not specific to this dataset or trained potential.

Generation composes hundreds of velocity evaluations, so we also compare the
two velocity fields end to end by sampling with each and measuring the FID of the generated samples.

\paragraph{Generation by relaxation.}
We run the Energy Matching sampler unchanged, replacing $v_{\rm FFN}$ with
$v_{\rm PCN}$ at every integration step, on our CIFAR-10 replication
checkpoints and on the authors' released ImageNet-32 checkpoints --- a scale
where training is beyond our compute budget but sampling is not.
Table~\ref{tab:ffn2pcn-fid} shows the FID matches feedforward inference in
every case. Within each row both runs start from identical noise,
so the differences reflect the velocity source rather than sampling variance;
the largest is $0.036$; App.~\ref{app:correspondence-measurement} shows samples
from both velocity sources side by side. A trained Energy Matching model can therefore be sampled at ImageNet-32 scale
by relaxation alone, with no measurable quality loss.

\begin{table}[tb]
\centering
\caption{Backprop-trained checkpoints sampled with feedforward autograd
velocities ($v_{\rm FFN}$) versus PCN relaxation velocities ($v_{\rm PCN}$),
at $\gamma = 10^{-4}$ (CIFAR-10) or $10^{-6}$ (ImageNet-32). Phases 1 and 2
are the Energy Matching training stages of Section~\ref{sec:background};
sampling seeds are matched within each row.}
\label{tab:ffn2pcn-fid}
\begin{tabular}{llcc}
\toprule
 & & \multicolumn{2}{c}{FID $\downarrow$} \\
\cmidrule(lr){3-4}
Dataset & Checkpoint & Feedforward & PCN relaxation \\
\midrule
CIFAR-10 & Phase 1, $\tau_s = 1.0$ & 6.42 & 6.42 \\
         & Phase 2, $\tau_s = 3.25$ & 3.54 & 3.55 \\
\midrule
ImageNet-32 & Phase 1, $\tau_s = 1.0$ & 8.33 & 8.33 \\
            & Phase 2, $\tau_s = 2.5$ & 6.75 & 6.78 \\
\bottomrule
\end{tabular}
\end{table}

\subsection{Equivalence at training}\label{subsec:equivalence-at-training}

We now ask what training on the PCN requires, comparing the
updates it produces --- by EP, $\Delta\theta_{\rm EP}$, and by the
implicit-function solve of Section~\ref{subsec:training},
$\Delta\theta_{\rm IFT}$ --- against the backpropagation gradient
$\Delta\theta_{\rm BP}$ of the same feedforward model on the same batch. The
optimiser rescales the update magnitude, so we measure only its direction, as a cosine
similarity with $\Delta\theta_{\rm BP}$. We relax the free phase to
convergence throughout (App.~\ref{app:training-fidelity}).

\paragraph{The correspondence itself is not the constraint.}
Where EP estimates the parameter gradient by a finite difference of two nudged
equilibria, the implicit solve obtains it directly at equilibrium, so
its only error is the $O(\gamma)$ correspondence bias of
Section~\ref{subsec:equivalence-at-inference}. Even in TF32
$\cos(\Delta\theta_{\rm IFT}, \Delta\theta_{\rm BP})$ is exact to four decimal
places from $\gamma = 10^{-7}$ to $10^{-2}$, softening only to $0.998$ at
$\gamma = 10^{-1}$, and is equally insensitive to the number of
conjugate-gradient steps (App.~\ref{app:training-fidelity}). 
Everything that
follows is therefore a test of EP's finite-difference estimator rather than of
the correspondence.

\paragraph{$\gamma$ and $\beta$ set a window together.}
Once the free phase has converged, the fidelity of $\Delta\theta_{\rm EP}$ to
$\Delta\theta_{\rm BP}$ depends on the size of the
displacement the nudge produces, which $\gamma$ and $\beta$ control jointly:
the two nudged equilibria differ by $O(\gamma\beta)$
(Section~\ref{subsec:linearised-nudge}), so it is their product that matters.
Table~\ref{tab:gamma-beta} (App.~\ref{app:training-fidelity}) gives
$\cos(\Delta\theta_{\rm EP}, \Delta\theta_{\rm BP})$ on a $\gamma$--$\beta$
grid, and the usable region is a diagonal band, where each decade lost in $\gamma$ is recovered by roughly a decade
gained in $\beta$: $\gamma = 3\times10^{-5}$ needs $\beta \geq 10$, where
$\gamma = 10^{-3}$ is clean from $\beta = 0.3$. The two edges have different
causes. Below the band the displacement is too small to survive TF32, and
float64 rescues it completely --- $1.0000$ against TF32's $0.77$ at the
operating $\gamma = 10^{-4}$ and $\beta = 0.3$ (Figure~\ref{fig:train-beta}). Above it the nudged equilibria are displaced too far for a linear finite difference to estimate the derivative well; float64 does not rescue that
edge, and since the implicit solve carries the same $O(\gamma)$ bias yet
degrades only to $0.998$ there, the failure belongs to the nudge rather than to
the correspondence or the arithmetic. The band therefore closes at both ends,
with no $\beta$ reaching $0.99$ by $\gamma = 10^{-1}$; our phase-1 operating point,
$\gamma = 10^{-4}$ with $\beta = 30$, sits comfortably inside it at $0.9998$. The window width is due to the linearised nudge: sweeping $\beta$ alone at the operating $\gamma$ (App.~\ref{app:training-fidelity}), the
linear nudge holds $\cos \geq 0.99$ from $\beta = 3$ to $10^3$, where the
quadratic nudge only does so at $\beta = 1$ and nowhere above.

\section{CIFAR-10 generation by backprop-free training}
\label{sec:cifar}

The previous section established both correspondences at fixed weights. We now
train with them end to end, over $10^5$ steps, and measure the generation
quality of the resulting model.

We train three arms: the feedforward Energy Matching model by
backpropagation, and the PCN at $\gamma = 10^{-4}$ by the implicit solve and
by EP with the linearised nudge. The architecture, optimiser, schedule, batch
size, EMA decay and step budget are shared; initialisation and the random
draws of training are not. The EP arm's phase-2 differs further: a
smaller $\gamma$, a stiffer tether $\lambda$ and a larger $\beta$, starting from its
$100$k phase-1 checkpoint rather than $145$k. Relaxation settings are in
Table~\ref{tab:hyperparams} and the rest of the recipe in
App.~\ref{app:experimental-implementation}.

\paragraph{Backprop-free training matches backpropagation.}
Table~\ref{tab:cifar-fid} gives the three arms. After phase 1, backpropagation
reaches an FID of $6.42$ at $145$k steps and the implicit solve $6.70$, while
the EP arm reaches $6.23$ at the $100$k checkpoint its contrastive phase starts
from. These span $0.47$, more than the $\sim 0.2$ below which we cannot resolve two FID evaluations (App.~\ref{app:experimental-implementation}), which is unsurprising for runs
that share neither an initialisation nor a data realisation. The contrastive phase then brings all three together: $3.54$ for
backpropagation, $3.64$ for the implicit solve, and $3.64$ for EP --- the one
estimator of the three that a neuromorphic substrate could run. The PCN is
therefore trainable end to end without backpropagation in the parameter updates of either phase. Both PCN
arms do draw their phase-2 negatives with a feedforward twin carrying the
same weights, since relaxing them was beyond our compute budget; no parameter
gradient passes through that twin, and
Section~\ref{subsec:equivalence-at-inference} shows the two velocity fields
agree (App.~\ref{app:experimental-implementation}).

The upper block of Table~\ref{tab:cifar-fid} places these numbers among
conventionally trained models. Our replication reaches $3.54$ against the
published $3.34$, a difference comparable to the repeat spread of the
measurement itself (App.~\ref{app:experimental-implementation}); on that scale
the backprop-free arms sit within the same band as recent diffusion and flow-matching
models, and well ahead of prior energy-based ones. To our knowledge this is the first CIFAR-10 generation result for a model
trained by Equilibrium Propagation.

    \begin{table}[tb]
    \centering
    \caption{Unconditional CIFAR-10 FID, with our models after both training phases at each
    checkpoint's optimal $\tau_s$. The upper block is taken from the cited
    papers; $^{*}$ marks reproductions by
    \citet{balcerakEnergyMatchingUnifying2025}. $^{\ddagger}$Phase-2 negatives drawn with a feedforward twin (see text). Protocol and per-arm differences: App.~\ref{app:experimental-implementation},
    App.~\ref{app:hardware-arch}.}
    \label{tab:cifar-fid}
    \begin{tabular}{llcc}
    \toprule
    Method & Gradient & Params & FID $\downarrow$ \\
    \midrule
    ImprovedCD \citep{duImprovedContrastiveDivergence2021} & backprop & --- & 25.1 \\
    SDiT (spiking) \citep{yangSDiTScalableEfficient2026} & backprop & 77M & 11.34 \\
    CLEL-large \citep{leeGuidingEnergybasedModels2022} & backprop & 31M & 8.61 \\
    OT-CFM$^{*}$ \citep{tongImprovingGeneralizingFlowbased2024} & backprop & 37M & 4.04 \\
    DDPM++ \citep{kimSoftTruncationUniversal2022} & backprop & 62M & 3.45 \\
    Energy Matching \citep{balcerakEnergyMatchingUnifying2025} & backprop & 50M & 3.34 \\
    \midrule
    Energy Matching (our replication) & backprop & 50M & 3.54 ($\tau_s = 3.25$) \\
    PotentialEP & implicit solve & 50M & 3.64 ($\tau_s = 3.25$)$^{\ddagger}$ \\
    PotentialEP & EP & 50M & 3.64 ($\tau_s = 3.25$)$^{\ddagger}$ \\
    \midrule
    Energy Matching, norm- and attention-free (\S\ref{subsec:hardware}) & backprop & 78M & 3.51 ($\tau_s = 5.0$) \\
    \bottomrule
    \end{tabular}
    \end{table}

\subsection{Hardware-plausible architectures}
\label{subsec:hardware}

PotentialEP is neuromorphically compatible as an algorithm, but the energy it
trains is only as neuromorphically implementable as the architecture defining it. Two
components we inherit from Energy Matching are not easily realised on an
analog substrate: group normalisation, which couples every channel at a layer
through a shared statistic, and attention, whose pairwise products are not a
fixed weighted sum. We therefore ask how much generation quality is lost when both are
removed.

Replacing them with weight standardisation, which normalises each filter's
weights rather than its activations, residual scaling by
$1/\sqrt{\text{depth}}$, and a flat linear head in place of the ViT
(App.~\ref{app:hardware-arch}) leaves an architecture with no operation beyond
fixed weighted sums and pointwise nonlinearities --- both native to analog
crossbar hardware, the sums through Ohm's and Kirchhoff's laws
\citep{ielminiInmemoryComputingResistive2018,
sebastianMemoryDevicesApplications2020}. We train this arm by backpropagation, our compute budget not extending to a
second full EP run, and it reaches an FID of $3.51$ against the full
architecture's $3.54$: the two are indistinguishable at the resolution of the
measurement. Generation competitive with the diffusion and flow-matching baselines of Table~\ref{tab:cifar-fid} therefore does not require either component. This
architecture is wider ($78$M against $50$M) and used a different schedule and
a longer first phase (App.~\ref{app:hardware-arch}), so we take it as an
existence proof and leave a cheaper recipe to future work.

\section{Related work}
\label{sec:related}

\paragraph{Predictive coding.} Generative predictive coding typically places the image at
the bottom of a hierarchy of predictions: Monte Carlo predictive coding samples
it by adding Langevin noise to inference, at MNIST scale
\citep{oliviersLearningProbabilityDistributions2024}, and Langevin predictive
coding samples a latent code the same way and decodes it, reaching CIFAR-10 at
FID $113$ \citep{zahidSampleYouInfer2023}. In both, the image is the predicted layer of a prediction-error likelihood. In
ours it is the input of a network whose scalar output is an explicit potential,
and the force on it is that potential's gradient.

\paragraph{Generation by energy descent.}
Energy Matching \citep{balcerakEnergyMatchingUnifying2025}, which we build on,
generates by descent on an explicit scalar potential. Equilibrium Matching
\citep{wangEquilibriumMatchingGenerative2025} learns a time-invariant gradient
field, in its best-performing form without an explicit energy, and
\citet{rodriguezGenerativeAssociativeMemory2026} train it on an Energy
Transformer, whose blocks relax latent tokens on a Hopfield-type energy,
reaching class-conditional FID $28.56$ on CIFAR-10. There the image-space update is a learned function of the relaxed tokens
rather than the gradient of any energy in $x$; in ours it is the gradient of
$\gamma^{-1}E_{\rm PCN} = V_\theta + O(\gamma)$. Each of these works trains by
backpropagation.

\paragraph{Physical and neuromorphic generation.}
A parallel line generates with physical substrates. Simulated Kuramoto
oscillators \citep{IntroducingUn0Generating} and continuous spins descending a
learned quadratic energy \citep{sunConditionalDynamicalSystems2026} evolve a
latent state from a random start and render it through a small convolutional
decoder, reaching class-conditional CIFAR-10 FID $8.76$ and clean-FID $9.71$
respectively, the former also at ImageNet $64\times64$. Optical systems have
implemented a diffusion model's denoiser \citep{oguzOpticalDiffusion2024} and a
single-pass generator \citep{chenOpticalGenerativeModels2025}, demonstrated in
hardware at MNIST scale, and spiking diffusion models target event-driven
neuromorphic chips \citep{caoSpikingDenoisingDiffusion2024}. All of these are
trained by backpropagation. \citet{boschLocalLearningRules2026} generate
$28\times28$ MNIST zeros and ones with a local rule, on an energy with no
hidden units. Ours relaxes in image space with no decoder, and uses
Equilibrium Propagation to obtain local updates for an energy with hidden
layers.

\section{Discussion and limitations}
\label{sec:discussion}
\paragraph{Simulation cost.}
On a GPU, an EP training step costs about sixteen times a backpropagation
step, and an implicit-solve step about five, with the relaxation budgets
dominating. A substrate that settles physically would not pay that cost, but
its speed, energy and settling accuracy remain unmeasured: everything here is
simulated.

\paragraph{Locality and hardware realisation.}
The relaxation is local in the sense EP requires: every quantity a weight
update needs is present at its two ends. The nodes of our training runs are,
however, architecture blocks
rather than single weight matrices, chosen to reduce simulation cost, so a
literal implementation would need circuitry to carry the error through each
block's internal layers. One node per weight matrix removes this, and left the training gradient
unchanged in the one checkpoint we measured (App.~\ref{app:dag-pcn}). Analog EP also obtains the backward drive $J^\top e$ from a feedback pathway
programmed to match the forward one; any drift between the two leaves
dynamics that descend no energy at
all~\citep{kendallTrainingEndtoEndAnalog2020}.

\paragraph{Scope of the evidence.}
App.~\ref{app:scope} collects the scope of each claim. Each arm is a single run, so training variance is unknown, and the small FID
differences do not establish a ranking. We demonstrate EP training on
CIFAR-10 only; at ImageNet-32 we verify the inference correspondence on the
authors' released checkpoints, but did not train. The hardware-plausible
architecture was trained and sampled only as a feedforward network, so it has
not been demonstrated together with EP. The equilibria were computed by
noise-free digital relaxation in error coordinates; the local state dynamics
share their fixed points (App.~\ref{app:error-param}) but were not run to
equilibrium at this scale. Analog noise, finite bit depth and
forward--backward pathway mismatch remain untested.

\newpage

\subsection*{AI use statement}

We used a frontier large language model in three ways. For the paper's prose, we used
it iteratively at the sentence level --- for wording, concision and clarity
--- rather than to draft sections or generate content; the ideas, claims and
structure are the author's. For the code, we used it to help write and debug
the training, sampling and measurement implementations; every suggestion was
read and accepted or rejected manually, and the numerical results reported
here come from runs of the accepted code. For related work, we used it to
expand a literature search the author had already conducted; we checked every
claim we attribute to a cited work against that work directly. We did not use
it to generate research ideas or hypotheses, to analyse results, or to produce
figures. We have reviewed all AI-assisted work and take responsibility for the
final content of this paper, including its text, claims and artifacts.

\subsection*{Reproducibility statement}

The PCN construction and the correspondence it rests on are given in
Section~\ref{subsec:correspondence}, with the full derivation, its assumptions
and the order of every neglected term in App.~\ref{app:derivation}; the
extension to the skip-connected UNet is in App.~\ref{app:dag-pcn} and the
solver we relax with in App.~\ref{app:error-param}. The training construction,
its parameters and the contrastive phase are in Section~\ref{subsec:training}
and App.~\ref{app:training-details}. App.~\ref{app:experimental-implementation}
gives every training and sampling setting, the FID protocol, the measurement
resolution, and the implementation choices the PCN formulation requires but a
feedforward one does not; App.~\ref{app:hardware-arch} specifies the norm- and
attention-free architecture and its recipe. Code for training,
sampling and the correspondence measurements is available at
\url{https://github.com/alexgower/EnergyMatchingEqProp}.

\subsection*{Acknowledgements}
We thank Joachim Wabnig for supervision and feedback on drafts, and Avner Karasik for insightful discussions. This work was
supported by UKRI/EPSRC CASE Award under Grant 220191 in partnership with
Nokia UK Limited. This work was performed using resources provided by the
Cambridge Service for Data Driven Discovery (CSD3) operated by the University
of Cambridge Research Computing Service (\url{www.csd3.cam.ac.uk}), provided
by Dell EMC and Intel using Tier-2 funding from the Engineering and Physical
Sciences Research Council (capital grant EP/T022159/1), and DiRAC funding from
the Science and Technology Facilities Council (\url{www.dirac.ac.uk}).

\bibliography{iclr2027_conference}

@misc{scellierEquilibriumPropagationBridging2017,
	title = {Equilibrium {Propagation}: {Bridging} the {Gap} {Between} {Energy}-{Based} {Models} and {Backpropagation}},
	shorttitle = {Equilibrium {Propagation}},
	url = {http://arxiv.org/abs/1602.05179},
	urldate = {2024-10-29},
	publisher = {arXiv},
	author = {Scellier, Benjamin and Bengio, Yoshua},
	month = mar,
	year = {2017},
	note = {arXiv:1602.05179},
}

@misc{laborieuxScalingEquilibriumPropagation2020,
	title = {Scaling {Equilibrium} {Propagation} to {Deep} {ConvNets} by {Drastically} {Reducing} its {Gradient} {Estimator} {Bias}},
	url = {http://arxiv.org/abs/2006.03824},
	urldate = {2024-11-08},
	publisher = {arXiv},
	author = {Laborieux, Axel and Ernoult, Maxence and Scellier, Benjamin and Bengio, Yoshua and Grollier, Julie and Querlioz, Damien},
	month = jun,
	year = {2020},
	note = {arXiv:2006.03824},
}

@misc{meerschTrainingHopfieldVariational2023,
	title = {Training a {Hopfield} {Variational} {Autoencoder} with {Equilibrium} {Propagation}},
	url = {http://arxiv.org/abs/2311.15047},
	doi = {10.48550/arXiv.2311.15047},
	urldate = {2025-08-07},
	publisher = {arXiv},
	author = {Meersch, Tom Van Der and Deleu, Johannes and Demeester, Thomas},
	month = nov,
	year = {2023},
	note = {arXiv:2311.15047 [cs]},
}

@misc{kendallTrainingEndtoEndAnalog2020,
	title = {Training {End}-to-{End} {Analog} {Neural} {Networks} with {Equilibrium} {Propagation}},
	url = {http://arxiv.org/abs/2006.01981},
	doi = {10.48550/arXiv.2006.01981},
	urldate = {2025-08-12},
	publisher = {arXiv},
	author = {Kendall, Jack and Pantone, Ross and Manickavasagam, Kalpana and Bengio, Yoshua and Scellier, Benjamin},
	month = jun,
	year = {2020},
	note = {arXiv:2006.01981 [cs]},
}

@misc{lipmanFlowMatchingGenerative2023,
	title = {Flow {Matching} for {Generative} {Modeling}},
	url = {http://arxiv.org/abs/2210.02747},
	doi = {10.48550/arXiv.2210.02747},
	urldate = {2025-12-17},
	author = {Lipman, Yaron and Chen, Ricky T. Q. and Ben-Hamu, Heli and Nickel, Maximilian and Le, Matt},
	month = feb,
	year = {2023},
	note = {arXiv:2210.02747 [cs]},
}

@misc{balcerakEnergyMatchingUnifying2025,
	title = {Energy {Matching}: {Unifying} {Flow} {Matching} and {Energy}-{Based} {Models} for {Generative} {Modeling}},
	shorttitle = {Energy {Matching}},
	url = {http://arxiv.org/abs/2504.10612},
	doi = {10.48550/arXiv.2504.10612},
	urldate = {2026-02-10},
	publisher = {arXiv},
	author = {Balcerak, Michal and Amiranashvili, Tamaz and Terpin, Antonio and Shit, Suprosanna and Bogensperger, Lea and Kaltenbach, Sebastian and Koumoutsakos, Petros and Menze, Bjoern},
	month = oct,
	year = {2025},
	note = {arXiv:2504.10612 [cs]},
}

@misc{tongImprovingGeneralizingFlowbased2024,
	title = {Improving and generalizing flow-based generative models with minibatch optimal transport},
	url = {http://arxiv.org/abs/2302.00482},
	doi = {10.48550/arXiv.2302.00482},
	urldate = {2026-04-08},
	publisher = {arXiv},
	author = {Tong, Alexander and Fatras, Kilian and Malkin, Nikolay and Huguet, Guillaume and Zhang, Yanlei and Rector-Brooks, Jarrid and Wolf, Guy and Bengio, Yoshua},
	month = mar,
	year = {2024},
	note = {arXiv:2302.00482 [cs]},
}

@misc{kerjanTrainingPredictiveCoding2026,
	title = {Training a {Predictive} {Coding} {Network} on {ImageNet} using {Equilibrium} {Propagation}},
	url = {http://arxiv.org/abs/2606.03584},
	doi = {10.48550/arXiv.2606.03584},
	urldate = {2026-06-26},
	publisher = {arXiv},
	author = {Kerjan, Tugdual and Høier, Rasmus and Scellier, Benjamin},
	month = jun,
	year = {2026},
	note = {arXiv:2606.03584 [cs.LG]},
}

@inproceedings{gowerFlowEqPropTrainingFlow2026,
	address = {New York, NY, USA},
	series = {{ICONS} '26},
	title = {{FlowEqProp}: {Training} {Flow} {Matching} {Models} with {Gradient} {Equilibrium} {Propagation}},
	isbn = {979-8-4007-2808-2},
	shorttitle = {{FlowEqProp}},
	url = {https://doi.org/10.1145/3822454.3822485},
	doi = {10.1145/3822454.3822485},
	urldate = {2026-07-16},
	booktitle = {Proceedings of the {International} {Conference} on {Neuromorphic} {Systems}},
	publisher = {Association for Computing Machinery},
	author = {Gower, Alex},
	month = aug,
	year = {2026},
	pages = {60--66},
}

@misc{qiaoMicroBatchTrainingBatchChannel2020,
	title = {Micro-{Batch} {Training} with {Batch}-{Channel} {Normalization} and {Weight} {Standardization}},
	url = {http://arxiv.org/abs/1903.10520},
	doi = {10.48550/arXiv.1903.10520},
	language = {en},
	urldate = {2026-09-16},
	publisher = {arXiv},
	author = {Qiao, Siyuan and Wang, Huiyu and Liu, Chenxi and Shen, Wei and Yuille, Alan},
	month = aug,
	year = {2020},
	note = {arXiv:1903.10520 [cs.CV]},
}

@misc{deBatchNormalizationBiases2020,
	title = {Batch {Normalization} {Biases} {Residual} {Blocks} {Towards} the {Identity} {Function} in {Deep} {Networks}},
	url = {http://arxiv.org/abs/2002.10444},
	doi = {10.48550/arXiv.2002.10444},
	urldate = {2026-09-16},
	publisher = {arXiv},
	author = {De, Soham and Smith, Samuel L.},
	month = dec,
	year = {2020},
	note = {arXiv:2002.10444 [cs.LG]},
}

@article{sebastianMemoryDevicesApplications2020,
	title = {Memory devices and applications for in-memory computing},
	volume = {15},
	copyright = {2020 Springer Nature Limited},
	issn = {1748-3395},
	url = {https://www.nature.com/articles/s41565-020-0655-z},
	doi = {10.1038/s41565-020-0655-z},
	language = {en},
	number = {7},
	urldate = {2026-09-17},
	journal = {Nature Nanotechnology},
	publisher = {Nature Publishing Group},
	author = {Sebastian, Abu and Le Gallo, Manuel and Khaddam-Aljameh, Riduan and Eleftheriou, Evangelos},
	month = jul,
	year = {2020},
	pages = {529--544},
}

@article{ielminiInmemoryComputingResistive2018,
	title = {In-memory computing with resistive switching devices},
	volume = {1},
	copyright = {2018 The Author(s)},
	issn = {2520-1131},
	url = {https://www.nature.com/articles/s41928-018-0092-2},
	doi = {10.1038/s41928-018-0092-2},
	language = {en},
	number = {6},
	urldate = {2026-09-17},
	journal = {Nature Electronics},
	publisher = {Nature Publishing Group},
	author = {Ielmini, Daniele and Wong, H.-S. Philip},
	month = jun,
	year = {2018},
	pages = {333--343},
}

@article{whittingtonApproximationErrorBackpropagation2017,
	title = {An {Approximation} of the {Error} {Backpropagation} {Algorithm} in a {Predictive} {Coding} {Network} with {Local} {Hebbian} {Synaptic} {Plasticity}},
	volume = {29},
	issn = {1530-888X},
	doi = {10.1162/NECO_a_00949},
	language = {eng},
	number = {5},
	journal = {Neural Computation},
	author = {Whittington, James C. R. and Bogacz, Rafal},
	month = may,
	year = {2017},
	pages = {1229--1262},
}

@article{oliviersLearningProbabilityDistributions2024,
	title = {Learning probability distributions of sensory inputs with {Monte} {Carlo} predictive coding},
	volume = {20},
	issn = {1553-7358},
	url = {https://journals.plos.org/ploscompbiol/article?id=10.1371/journal.pcbi.1012532},
	doi = {10.1371/journal.pcbi.1012532},
	language = {en},
	number = {10},
	urldate = {2026-09-17},
	journal = {PLOS Computational Biology},
	publisher = {Public Library of Science},
	author = {Oliviers, Gaspard and Bogacz, Rafal and Meulemans, Alexander},
	month = oct,
	year = {2024},
	pages = {e1012532},
}

@misc{zahidSampleYouInfer2023,
	title = {Sample as {You} {Infer}: {Predictive} {Coding} {With} {Langevin} {Dynamics}},
	shorttitle = {Sample as {You} {Infer}},
	url = {http://arxiv.org/abs/2311.13664},
	doi = {10.48550/arXiv.2311.13664},
	urldate = {2026-09-22},
	publisher = {arXiv},
	author = {Zahid, Umais and Guo, Qinghai and Fountas, Zafeirios},
	month = nov,
	year = {2023},
	note = {arXiv:2311.13664 [cs.LG]
version: 1},
}

@misc{sunConditionalDynamicalSystems2026,
	title = {Conditional {Dynamical} {Systems} for {Image} {Generation}},
	url = {http://arxiv.org/abs/2608.14961},
	doi = {10.48550/arXiv.2608.14961},
	urldate = {2026-09-22},
	publisher = {arXiv},
	author = {Sun, Lianlong and Liu, Chuan and Geng, Tong and Huang, Michael},
	month = aug,
	year = {2026},
	note = {arXiv:2608.14961 [cs.ET]},
}

@article{chenOpticalGenerativeModels2025,
	title = {Optical {Generative} {Models}},
	volume = {644},
	issn = {0028-0836, 1476-4687},
	url = {http://arxiv.org/abs/2410.17970},
	doi = {10.1038/s41586-025-09446-5},
	number = {8078},
	urldate = {2026-09-22},
	journal = {Nature},
	author = {Chen, Shiqi and Li, Yuhang and Chen, Hanlong and Ozcan, Aydogan},
	month = aug,
	year = {2025},
	note = {arXiv:2410.17970 [cs.NE]},
	pages = {903--911},
}

@article{millidgePredictiveCodingApproximates2022,
	title = {Predictive {Coding} {Approximates} {Backprop} {Along} {Arbitrary} {Computation} {Graphs}},
	volume = {34},
	issn = {0899-7667},
	url = {https://doi.org/10.1162/neco_a_01497},
	doi = {10.1162/neco_a_01497},
	number = {6},
	urldate = {2026-09-22},
	journal = {Neural Computation},
	author = {Millidge, Beren and Tschantz, Alexander and Buckley, Christopher L.},
	month = may,
	year = {2022},
	pages = {1329--1368},
}

@article{raoPredictiveCodingVisual1999,
	title = {Predictive coding in the visual cortex:  a functional interpretation of some extra-classical receptive-field effects},
	volume = {2},
	issn = {1546-1726},
	url = {https://doi.org/10.1038/4580},
	doi = {10.1038/4580},
	number = {1},
	journal = {Nature Neuroscience},
	author = {Rao, Rajesh P. N. and Ballard, Dana H.},
	month = jan,
	year = {1999},
	pages = {79--87},
}

@inproceedings{hoierDualPropagationAccelerating2023,
	title = {Dual {Propagation}: {Accelerating} {Contrastive} {Hebbian} {Learning} with {Dyadic} {Neurons}},
	issn = {2640-3498},
	shorttitle = {Dual {Propagation}},
	url = {https://proceedings.mlr.press/v202/hoier23a.html},
	language = {en},
	urldate = {2026-09-22},
	booktitle = {Proceedings of the 40th {International} {Conference} on {Machine} {Learning}},
	publisher = {PMLR},
	author = {Høier, Rasmus and Staudt, D. and Zach, Christopher},
	month = jul,
	year = {2023},
	pages = {13141--13156},
}

@misc{wangEquilibriumMatchingGenerative2025,
	title = {Equilibrium {Matching}: {Generative} {Modeling} with {Implicit} {Energy}-{Based} {Models}},
	shorttitle = {Equilibrium {Matching}},
	url = {http://arxiv.org/abs/2510.02300},
	doi = {10.48550/arXiv.2510.02300},
	urldate = {2026-09-22},
	publisher = {arXiv},
	author = {Wang, Runqian and Du, Yilun},
	month = oct,
	year = {2025},
	note = {arXiv:2510.02300 [cs.LG]},
}

@misc{IntroducingUn0Generating,
	title = {Introducing {Un}-0: {Generating} {Images} with {Coupled} {Oscillators} - {Unconventional} {AI}},
	url = {https://unconv.ai/blog/introducing-un-0-generating-images-with-coupled-oscillators/},
	urldate = {2026-09-22},
	author = {{Unconventional AI}},
	year = {2026},
}

@inproceedings{oguzOpticalDiffusion2024,
	title = {Optical diffusion models for image generation},
	volume = {37},
	url = {https://proceedings.neurips.cc/paper_files/paper/2024/file/6cb81234ab47027e991728ed7dd76735-Paper-Conference.pdf},
	doi = {10.52202/079017-1887},
	booktitle = {Advances in neural information processing systems},
	publisher = {Curran Associates, Inc.},
	author = {Oguz, Ilker and Dinc, Niyazi Ulas and Yildirim, Mustafa and Ke, Junjie and Yoo, Innfarn and Wang, Qifei and Yang, Feng and Moser, Christophe and Psaltis, Demetri},
	editor = {Globerson, A. and Mackey, L. and Belgrave, D. and Fan, A. and Paquet, U. and Tomczak, J. and Zhang, C.},
	year = {2024},
	pages = {59150--59173},
}

@misc{boschLocalLearningRules2026,
	title = {Local {Learning} {Rules} for {Out}-of-{Equilibrium} {Physical} {Generative} {Models}},
	url = {http://arxiv.org/abs/2506.19136},
	doi = {10.48550/arXiv.2506.19136},
	urldate = {2026-09-22},
	publisher = {arXiv},
	author = {Bösch, Cyrill and Roeder, Geoffrey and Serra-Garcia, Marc and Adams, Ryan P.},
	month = jul,
	year = {2026},
	note = {arXiv:2506.19136 [cs.LG]},
}

@inproceedings{caoSpikingDenoisingDiffusion2024,
	address = {Waikoloa, HI, USA},
	title = {Spiking {Denoising} {Diffusion} {Probabilistic} {Models}},
	copyright = {https://doi.org/10.15223/policy-029},
	isbn = {979-8-3503-1892-0},
	url = {https://ieeexplore.ieee.org/document/10484223/},
	doi = {10.1109/WACV57701.2024.00484},
	language = {en},
	urldate = {2026-09-22},
	booktitle = {2024 {IEEE}/{CVF} {Winter} {Conference} on {Applications} of {Computer} {Vision} ({WACV})},
	publisher = {IEEE},
	author = {Cao, Jiahang and Wang, Ziqing and Guo, Hanzhong and Cheng, Hao and Zhang, Qiang and Xu, Renjing},
	month = jan,
	year = {2024},
	pages = {4900--4909},
}

@article{yangSDiTScalableEfficient2026,
	title = {{SDiT}: {Scalable} and efficient spiking diffusion transformers for image generation},
	volume = {207},
	issn = {0167-8655},
	shorttitle = {{SDiT}},
	url = {https://doi.org/10.1016/j.patrec.2026.07.003},
	doi = {10.1016/j.patrec.2026.07.003},
	number = {C},
	urldate = {2026-09-22},
	journal = {Pattern Recogn. Lett.},
	author = {Yang, Shu and Ma, Hanzhi and Yu, Chengting and Wang, Aili and Fu, Pengqi and Li, Er-Ping},
	month = aug,
	year = {2026},
	pages = {266--273},
}

@inproceedings{kimSoftTruncationUniversal2022,
	title = {Soft {Truncation}: {A} {Universal} {Training} {Technique} of {Score}-based {Diffusion} {Model} for {High} {Precision} {Score} {Estimation}},
	issn = {2640-3498},
	shorttitle = {Soft {Truncation}},
	url = {https://proceedings.mlr.press/v162/kim22i.html},
	language = {en},
	urldate = {2026-09-22},
	booktitle = {Proceedings of the 39th {International} {Conference} on {Machine} {Learning}},
	publisher = {PMLR},
	author = {Kim, Dongjun and Shin, Seungjae and Song, Kyungwoo and Kang, Wanmo and Moon, Il-Chul},
	month = jun,
	year = {2022},
	pages = {11201--11228},
}

@inproceedings{strubellEnergyPolicyConsiderations2019,
	address = {Florence, Italy},
	title = {Energy and {Policy} {Considerations} for {Deep} {Learning} in {NLP}},
	url = {https://aclanthology.org/P19-1355/},
	doi = {10.18653/v1/P19-1355},
	urldate = {2026-09-22},
	booktitle = {Proceedings of the 57th {Annual} {Meeting} of the {Association} for {Computational} {Linguistics}},
	publisher = {Association for Computational Linguistics},
	author = {Strubell, Emma and Ganesh, Ananya and McCallum, Andrew},
	editor = {Korhonen, Anna and Traum, David and Màrquez, Lluís},
	month = jul,
	year = {2019},
	pages = {3645--3650},
}

@article{wrightDeepPhysicalNeural2022,
	title = {Deep physical neural networks trained with backpropagation},
	volume = {601},
	copyright = {2022 The Author(s)},
	issn = {1476-4687},
	url = {https://www.nature.com/articles/s41586-021-04223-6},
	doi = {10.1038/s41586-021-04223-6},
	language = {en},
	number = {7894},
	urldate = {2026-09-22},
	journal = {Nature},
	publisher = {Nature Publishing Group},
	author = {Wright, Logan G. and Onodera, Tatsuhiro and Stein, Martin M. and Wang, Tianyu and Schachter, Darren T. and Hu, Zoey and McMahon, Peter L.},
	month = jan,
	year = {2022},
	pages = {549--555},
}

@inproceedings{duImprovedContrastiveDivergence2021,
	series = {Proceedings of machine learning research},
	title = {Improved contrastive divergence training of energy-based models},
	volume = {139},
	url = {https://proceedings.mlr.press/v139/du21b.html},
	booktitle = {Proceedings of the 38th international conference on machine learning},
	publisher = {PMLR},
	author = {Du, Yilun and Li, Shuang and Tenenbaum, Joshua and Mordatch, Igor},
	editor = {Meila, Marina and Zhang, Tong},
	month = jul,
	year = {2021},
	pages = {2837--2848},
}

@misc{goemaereEPCOvercomingExponential2025,
	title = {{ePC}: {Fast} and {Deep} {Predictive} {Coding} in {Digital} {Simulation}},
	shorttitle = {{ePC}},
	url = {http://arxiv.org/abs/2505.20137},
	doi = {10.48550/arXiv.2505.20137},
	urldate = {2026-09-24},
	publisher = {arXiv},
	author = {Goemaere, Cédric and Oliviers, Gaspard and Bogacz, Rafal and Demeester, Thomas},
	month = jun,
	year = {2026},
	note = {arXiv:2505.20137 [cs.LG]},
}

@inproceedings{leeGuidingEnergybasedModels2022,
	title = {Guiding {Energy}-based {Models} via {Contrastive} {Latent} {Variables}},
	url = {https://openreview.net/forum?id=CZmHHj9MgkP},
	language = {en},
	urldate = {2026-09-24},
	booktitle = {{ICLR}},
	author = {Lee, Hankook and Jeong, Jongheon and Park, Sejun and Shin, Jinwoo},
	month = sep,
	year = {2022},
}

@inproceedings{luccioniPowerHungryProcessing2024,
	address = {New York, NY, USA},
	series = {{FAccT} '24},
	title = {Power {Hungry} {Processing}: {Watts} {Driving} the {Cost} of {AI} {Deployment}?},
	isbn = {979-8-4007-0450-5},
	shorttitle = {Power {Hungry} {Processing}},
	url = {https://dl.acm.org/doi/10.1145/3630106.3658542},
	doi = {10.1145/3630106.3658542},
	urldate = {2026-09-23},
	booktitle = {Proceedings of the 2024 {ACM} {Conference} on {Fairness}, {Accountability}, and {Transparency}},
	publisher = {Association for Computing Machinery},
	author = {Luccioni, Sasha and Jernite, Yacine and Strubell, Emma},
	month = jun,
	year = {2024},
	pages = {85--99},
}

@inproceedings{rodriguezGenerativeAssociativeMemory2026,
	title = {Generative {Associative} {Memory} via {Equilibrium} {Matching}},
	url = {https://openreview.net/forum?id=4zjhUaeeVh},
	language = {en},
	urldate = {2026-09-24},
	booktitle = {{ICLR}},
	author = {Rodriguez, Adrian and Hoover, Benjamin and Guo, Yunhui and Du, Yilun},
	month = mar,
	year = {2026},
}
\bibliographystyle{iclr2027_conference}

\appendix

\section{Derivation of the correspondence}\label{app:derivation}

This appendix fills in Proposition~\ref{prop:correspondence}: the assumptions,
the order of every neglected term, and where the $O(\gamma)$ error comes from.
Throughout, $\alpha$ is a fixed constant of the architecture and is absorbed
into the $O(\cdot)$ symbols, so ``$O(\gamma)$'' means $O(\alpha\gamma)$.

\subsection{The argument in outline}

Note first that without the injection the energy is minimised at $e = 0$, that
is, at the forward pass $h_k^{\rm FFN} = (f_k \circ \cdots \circ f_1)(x)$ for
every $k$. The injection
disturbs this. Stationarity in the output node gives
$\partial E/\partial h_N = -e_N + \gamma\alpha = 0$, so the top error is a
constant, $e_N^* = \gamma\alpha$, exactly. Stationarity in
an interior node gives $\partial E/\partial h_k = -e_k + J_{k+1}^{\top}
e_{k+1} = 0$ with $J_{k+1} = \partial f_{k+1}/\partial h_k$, so
\begin{equation}\label{eq:residual-chain}
    e_k^* = J_{k+1}^{\top} e_{k+1}^* ,
    \qquad\text{hence}\qquad
    e_1^* = \gamma\alpha\; J_2^{\top} J_3^{\top} \cdots J_N^{\top} ,
\end{equation}
with every Jacobian evaluated at $h^*$. Each error is therefore the derivative
of the output with respect to its own node, propagated down by transposed
Jacobians from a seed of $\gamma\alpha$ at the top --- the same recursion
backpropagation performs, seeded there by $\partial V/\partial V = 1$. Every
error is $O(\gamma)$, so from $h_k^* = f_k(h_{k-1}^*) - e_k^*$ each node $h_k^*$ sits
$O(\gamma)$ from the forward pass state $h_k^{\rm FFN}$ it would occupy at $\gamma = 0$.

The velocity needs the total derivative of $E_{\rm PCN}(x, h^*(x))$ with
respect to $x$. Only the first error term depends on $x$ explicitly, contributing
$\partial E/\partial x = J_1^{\top} e_1$; the dependence through $h^*(x)$ adds
$\sum_k (\partial E/\partial h_k)\big|_{h^*}\,(\mathrm{d}h_k^*/\mathrm{d}x)$,
which vanishes term by term because $h^*$ is stationary. Hence
\begin{equation}\label{eq:envelope}
    \nabla_x E_{\rm PCN}\big(x, h^*(x)\big) = J_1^{\top} e_1^*
    = \gamma\alpha\; J_1^{\top} J_2^{\top} \cdots J_N^{\top}
    = \gamma\,\alpha \nabla_x F_\theta(x) + O(\gamma^2),
\end{equation}
the last step evaluating each Jacobian at $h^{\rm FFN}$ rather than at $h^*$,
which costs $O(\gamma)$ each, and recognising the resulting product as the
chain rule for $\nabla_x F_\theta$. Dividing by $\gamma$ and using
$V_\theta = \alpha F_\theta$ gives the first claim. For the second, evaluate
$E_{\rm PCN}$ at $h^*$: the error terms are $O(\gamma^2)$ by
\eqref{eq:residual-chain}, and $o^* = f_N(h_{N-1}^*) - \gamma\alpha =
F_\theta(x) + O(\gamma)$, so the injection term contributes
$\gamma V_\theta(x) + O(\gamma^2)$.

\subsection{The argument in detail}

This section repeats the steps above with the assumptions stated and every
neglected term accounted for.

\paragraph{Setting.} $E_{\rm PCN}$ is \eqref{eq:pcn-energy} on a chain of $N$
nodes with $h_0 = x$ and $h_N = o$ scalar. Each $f_k$ is $C^2$ in its input and in its parameters $\theta$, with bounded
first and second derivatives on the relevant region (SiLU, tanh; ReLU almost
everywhere). The differentiability in $\theta$ that
Section~\ref{subsec:training} requires needs $C^4$, which smooth activations
have. Write
$h_k^{\rm FFN} = (f_k \circ \cdots \circ f_1)(x)$ for the forward-pass state
at node $k$, with $h_0^{\rm FFN} = x$ and $h_N^{\rm FFN} = F_\theta(x)$, and
$J_k(h) = \partial f_k / \partial h$ for the Jacobian of layer $k$ at its
input. The chain rule gives $\nabla_x F_\theta(x) = J_1^{\top} J_2^{\top}
\cdots J_N^{\top}$, where each $J_k$ is evaluated at the input layer $k$
receives on the forward pass, namely $h_{k-1}^{\rm FFN}$; note $J_N$ is a row
vector, so the product is a column of dimension $\dim x$.

\paragraph{Existence of the equilibrium.} At $\gamma = 0$, $E_{\rm PCN} \geq 0$ with a unique minimum at $e = 0$, i.e.\
$h_k = f_k(h_{k-1})$ for every $k$, which by induction from $h_0 = x$ is the
forward pass $h_k = h_k^{\rm FFN}$. In the error
coordinates of App.~\ref{app:error-param} the Hessian there is the identity, so
by the implicit function theorem the minimiser continues smoothly to small
$\gamma > 0$, and depends smoothly on $\theta$ and $\gamma$
jointly; $h^*(x)$ denotes that branch. Physically it is the state the
hidden layers settle to when they relax faster than $x$ moves.

\paragraph{Step 1: the top error.} $h_N$ enters $E_{\rm PCN}$ through
$\tfrac12 (f_N(h_{N-1}) - h_N)^2$ and $\gamma\alpha h_N$, so
$\partial E/\partial h_N = -e_N + \gamma\alpha$, and stationarity gives
\begin{equation}
    e_N^* = \gamma\alpha \quad\text{exactly.}
\end{equation}
This is a constant independent of $x$ and $\theta$: the injection seeds the
backward pass with a fixed scalar.

\paragraph{Step 2: interior errors backpropagate.} For $1 \leq k < N$, $h_k$
enters its own error term and the next node's, giving
$\partial E/\partial h_k = -e_k + J_{k+1}(h_k)^{\top} e_{k+1}$. Stationarity
gives $e_k^* = J_{k+1}(h_k^*)^{\top} e_{k+1}^*$, and iterating from the top,
\begin{equation}\label{eq:app-residual-chain}
    e_k^* = \gamma\alpha\; J_{k+1}(h_k^*)^{\top} \cdots J_N(h_{N-1}^*)^{\top},
    \qquad k = 1, \dots, N-1 ,
\end{equation}
exact, with every Jacobian at the equilibrium state. Bounded Jacobians make
every $e_k^* = O(\gamma)$.

\paragraph{Step 3: states sit $O(\gamma)$ from the forward pass.} From
$h_k^* = f_k(h_{k-1}^*) - e_k^*$ and $h_0 = x$, each node's displacement from
the forward pass is its own $O(\gamma)$ error plus whatever its predecessor's
displacement becomes after passing through $f_k$. Bounded first derivatives
make $f_k$ Lipschitz with constant $L_k = \sup\|J_k\|$, so the displacement
compounds as
$\|h_k^* - h_k^{\rm FFN}\| \leq L_k \|h_{k-1}^* - h_{k-1}^{\rm FFN}\| +
\|e_k^*\|$ rather than growing without control, so $h_k^* = h_k^{\rm FFN} +
O(\gamma)$ at every depth. Bounded second derivatives then carry that displacement into the Jacobians.
Since $J_{k+1}$ is a function of the point at which it is evaluated, Taylor
expanding about $h_k^{\rm FFN}$ gives
\begin{equation}
    J_{k+1}(h_k^*) = J_{k+1}(h_k^{\rm FFN})
    \;+\; \nabla^2 f_{k+1}\big(h_k^{\rm FFN}\big)\,\big(h_k^* - h_k^{\rm FFN}\big)
    \;+\; \dots
\end{equation}
and the displacement in the second term is $O(\gamma)$ by the above, so
$J_{k+1}(h_k^*) = J_{k+1}(h_k^{\rm FFN}) + O(\gamma)$.

\paragraph{Step 4: the visible gradient.} $x = h_0$ appears in $E_{\rm PCN}$
only through $\tfrac12\|f_1(x) - h_1\|^2$, so at fixed $h$,
$\partial_x E_{\rm PCN} = J_1(x)^{\top} e_1$. The total derivative of $E_{\rm PCN}(x, h^*(x))$ is
$\partial_x E\big|_{h^*} + \sum_k (\partial_{h_k} E)\big|_{h^*}\,
(\mathrm{d}h_k^*/\mathrm{d}x)$, and the second sum vanishes because $h^*$ is
a stationary point of $E_{\rm PCN}$ in $h$. Hence
\begin{equation}
    \nabla_x E_{\rm PCN}\big(x, h^*(x)\big)
    = J_1(x)^{\top} e_1^*
    = \gamma\alpha\; J_1(x)^{\top} J_2(h_1^*)^{\top} \cdots J_N(h_{N-1}^*)^{\top}.
\end{equation}
Replacing each $J_{k}(h_{k-1}^*)$ by $J_k(h_{k-1}^{\rm FFN}) + O(\gamma)$ and
collecting, the product is $\nabla_x F_\theta(x) + O(\gamma)$, so
\begin{equation}
    \nabla_x E_{\rm PCN}\big(x, h^*(x)\big)
    = \gamma\alpha \nabla_x F_\theta(x) + O(\gamma^2)
    = \gamma \nabla_x V_\theta(x) + O(\gamma^2),
\end{equation}
and $v_{\rm PCN} = -\gamma^{-1} \nabla_x E_{\rm PCN}(x, h^*) = -\nabla_x
V_\theta(x) + O(\gamma)$.

\paragraph{Step 5: the energy value.} At equilibrium the error terms contribute
$\sum_k \tfrac12 \|e_k^*\|^2 = O(\gamma^2)$. For the injection term, Step 1
gives $e_N^* = \gamma\alpha$ and Step 3 gives $h_{N-1}^* = h_{N-1}^{\rm FFN} +
O(\gamma)$, so $o^* = f_N(h_{N-1}^*) - e_N^* = F_\theta(x) + O(\gamma)$ and
$E_{\rm PCN}(x, h^*) = \gamma\alpha F_\theta(x) + O(\gamma^2) = \gamma
V_\theta(x) + O(\gamma^2)$.

\paragraph{Where the error lives.} The stationarity conditions of Steps 1 and 2
and the envelope identity of Step 4 are exact, and $J_1$ is evaluated at $x$ itself, which is held fixed and carries no displacement. The only approximation is Step 3's
replacement of the interior Jacobians $J_2, \ldots, J_N$ by their values on the
forward pass, so the $O(\gamma)$ correction to $v_{\rm PCN}$ comes entirely
from the interior states being displaced by their own errors: it vanishes for
linear layers and is bounded by their second derivatives. The output node's own
displacement plays no part --- it has no children, and $e_N^*$ enters
everything downstream only through the exact constant $\gamma\alpha$.
Section~\ref{subsec:equivalence-at-inference} measures the correction as linear
in $\gamma$ over six decades, as this predicts.

\paragraph{Directed acyclic graphs.} On the skip-connected UNet of
App.~\ref{app:dag-pcn}, a node may have several parents and several children.
Step 2 becomes $e_j^* = \sum_{c \in \mathrm{ch}(j)} J_{c,j}^{\top} e_c^*$ with
$J_{c,j} = \partial f_c / \partial h_j$, which is exactly the chain rule's
multivariable form, and Step 4 is unchanged: every node fed by $x$ contributes
$J_{c,0}^{\top} e_c^*$ to the visible gradient. The rest of the argument goes
through verbatim.

\paragraph{Differentiability in $\theta$.} Proposition~\ref{prop:correspondence}
bounds the gap between $v_{\rm PCN}$ and $-\nabla_x V_\theta$ at fixed
$\theta$. The gradient correspondence of Section~\ref{subsec:training}, on which
training rests, needs more: that the gap's $\theta$-derivative is $O(\gamma)$
too, which does not follow from a pointwise bound alone. It follows instead
from smoothness, in three steps.

First, the equilibrium is as smooth as the layers. Applying the implicit
function theorem to the spring-clamped system \eqref{eq:free-energy} in the
joint variables $(x, e)$, whose Hessian at $\gamma = 0$ is $\lambda I \oplus
I$ and hence invertible, the free equilibrium $(x^*, h^*)$ is a $C^k$ function
of $(\theta, \gamma)$ whenever the layers are $C^{k+1}$.

Second, the velocity is finite at $\gamma = 0$. Every error vanishes there,
so $\nabla_x E_{\rm PCN}(x^*, h^*)$ vanishes and $x^* = x_t$; a $C^k$ function
that vanishes at $\gamma = 0$, divided by $\gamma$, is $C^{k-1}$, so
$v^* = -\gamma^{-1}\nabla_x E_{\rm PCN}$ is $C^{k-1}$ in $(\theta, \gamma)$
including at $\gamma = 0$, where by Step 4 its value is
$-\nabla_x V_\theta(x_t)$. Hence $\mathcal{L}_{\rm PCN}(\theta, 0) =
\mathcal{L}(\theta)$.

Third, the mean value theorem in $\gamma$: if $\nabla_\theta
\mathcal{L}_{\rm PCN}$ has a bounded $\gamma$-derivative, its change from
$\gamma = 0$ is at most that bound times $\gamma$. That derivative is a mixed
second derivative of $\mathcal{L}_{\rm PCN}$, so $\mathcal{L}_{\rm PCN}$ ---
and hence $v^*$, of which it is a smooth function --- must be jointly $C^2$.
Tracing back through the two derivatives lost above, one to the implicit
function theorem and one to the division by $\gamma$, this holds for layers
with four continuous derivatives, which smooth activations such as SiLU and tanh have. Then
$\nabla_\theta\mathcal{L}_{\rm PCN}(\theta, \gamma) = \nabla_\theta
\mathcal{L}(\theta) + O(\gamma)$, which is the correspondence.

\section{The training construction in detail}
\label{app:training-details}

This appendix fills in the construction of Section~\ref{subsec:training}.

$\mathcal{L}_{\rm PCN}$ differs from
$\mathcal{L}$ in two ways, both $O(\gamma)$. First, the
correspondence is inexact at finite $\gamma$: $v_{\rm PCN} = -\nabla_x
V_\theta + O(\gamma)$ by Proposition~\ref{prop:correspondence}. Second, $v^* =
v_{\rm PCN}(x^*)$ is the velocity at the relaxed state $x^*$ rather than at
$x_t$, an evaluation point displaced by $x^* - x_t = \gamma v^*/\lambda$. Keeping that displacement small is the tether $\lambda$'s job, and it has less
work to do here than in FlowEqProp \citep{gowerFlowEqPropTrainingFlow2026}, where GradEP was originally introduced: there the velocity converted
to an extension with a factor $O(1/\lambda)$, so a close evaluation point
required $\lambda \gg 1$, whereas \eqref{eq:velocity-readout} converts with
$\gamma/\lambda$ so the requirement is only $\gamma/\lambda \ll 1$. The
smallness of $\gamma$ that the correspondence already demands therefore
supplies this constraint too, leaving $\lambda$ free to be chosen on other
grounds (see below).

\paragraph{$\gamma$ and $\beta$.} The injection strength $\gamma$ plays for $V$ the
role $\beta$ plays for the cost $C$ in Equilibrium Propagation: a small linear
term on one node whose stationary response, divided by that term's strength,
is a gradient. As in EP, the division is a constant gain on the readout, which
costs nothing on an analog substrate.

\paragraph{Sweep budgets.} The free phase relaxes for $T_{\rm free}$ sweeps
and each nudged phase for $T_{\rm nudge}$ sweeps from the free equilibrium,
with $h$ equilibrated by $K_h$ inner sweeps per step of $x$ in both.

\paragraph{Parameters.}
Five quantities appear in the training construction besides $\gamma$. The spring stiffness $\lambda$ is not tuned: with the visible state updated as
$x \leftarrow x - \mathrm{d}t_x\,\nabla_x E_{\rm free}$, the spring extension
contracts by $(1 - \mathrm{d}t_x\,\lambda)$ per sweep, up to the $O(\gamma)$
curvature of $E_{\rm PCN}$ in $x$ (App.~\ref{app:linearised-nudge}), so
$\lambda\,\mathrm{d}t_x = 1$ settles it in a single step, exactly as the error
parameterisation settles $h$ in one sweep; larger $\lambda\,\mathrm{d}t_x$
overshoots. Phase 1 uses
$\lambda = 1$, $\mathrm{d}t_x = 1$ and phase 2 uses $\lambda = 2$,
$\mathrm{d}t_x = 0.5$ (App.~\ref{app:experimental-implementation}). The nudge strength $\beta$ is the
one parameter that must be chosen: it sets the displacement between the two
nudged equilibria, which must be large enough to resolve in finite-precision
arithmetic and small enough for \eqref{eq:ep-update} to hold, a window
Section~\ref{subsec:linearised-nudge} derives and
Section~\ref{subsec:equivalence-at-training} measures. The budgets
$T_{\rm free}$, $T_{\rm nudge}$ and $K_h$ are the sweep counts of the three
relaxations; the last two are satisfied by a single sweep for the same deadbeat
reason as $\lambda$, so the cost of an EP update is the free-phase
equilibration $T_{\rm free}$ alone.

\paragraph{Contrastive divergence.}
Energy Matching's second phase adds to the flow-matching loss a contrastive
term, $\nabla_\theta V(x^+) - \nabla_\theta V(x^-)$ between data and model
samples. On the PCN this needs no nudge at all. Writing $V = \gamma^{-1} E_{\rm PCN}(x, h^*(x;\theta))$, the parameter derivative of
$E_{\rm PCN}$ is $\partial_\theta E_{\rm PCN}|_{h^*} + \sum_k \partial_{h_k}
E_{\rm PCN}|_{h^*}\,(\mathrm{d}h_k^*/\mathrm{d}\theta)$, whose second term
vanishes at equilibrium. What remains is local: for a parameter in layer $k$,
$\partial E_{\rm PCN}/\partial\theta_k = e_k^{\top}\,\partial f_k /
\partial\theta_k$, involving only the error at node $k$ and the state feeding
it. The contrastive gradient is therefore the difference of $\gamma^{-1}
\partial_\theta E_{\rm PCN}|_{h^*}$ between the hidden equilibria reached with
$x$ clamped at $x^+$ and at $x^-$ --- a contrastive Hebbian rule.

\section{Finite-difference accuracy of the linearised nudge}
\label{app:linearised-nudge}

Write $d = x^* - x_t$ and
$d_\beta = x^{*\beta} - x_t$ for the free and nudged extensions, so that
$\nabla_x E_{\rm PCN}(x^*) = -\lambda d$ is the force balance at the free
equilibrium. Moving $x$ changes $\nabla_x E_{\rm PCN}$ by
$\nabla_x^2 E_{\rm PCN}$ times the displacement, and $\nabla_x^2 E_{\rm PCN} =
O(\gamma)$: the gradient $\nabla_x E_{\rm PCN} = J_1^{\top} e_1$ is $O(\gamma)$
in size (App.~\ref{app:derivation}, Step 2), and the distance in $x$ over which
it changes appreciably is set by the $O(1)$ weights and nonlinearities rather
than by $\gamma$, so each of its $x$-derivatives is $O(\gamma)$ too. The
tether's Hessian is $\nabla_x^2\,\tfrac{\lambda}{2}\|x - x_t\|^2 = \lambda I$,
so across the same displacement $\nabla_x E_{\rm PCN}$ varies by a fraction
$O(\gamma/\lambda)$ of what the tether does: we treat $\nabla_x E_{\rm PCN}$ as
constant at its free-phase value $-\lambda d$ throughout the nudge, while
keeping the full variation of the tether and nudge terms.

Stationarity of $E_{\rm free} + \beta' C$ in $x$ then sets three gradient terms
against each other, and the nudge term differs between the two costs.

\emph{Quadratic cost.} With $v(x) = \tfrac{\lambda}{\gamma}(x - x_t)$,
\begin{equation}
    \beta' C(x) = \beta\Big(\tfrac{\gamma}{\lambda}\Big)^2
        \tfrac12 \big\|\tfrac{\lambda}{\gamma}(x - x_t) - u\big\|^2 ,
    \qquad
    \nabla_x\,\beta' C = \beta\Big(x - x_t - \tfrac{\gamma}{\lambda}u\Big)
        = \beta\Big(d_\beta - \tfrac{\gamma}{\lambda}u\Big) ,
\end{equation}
so stationarity reads
\begin{equation}
    \underbrace{-\lambda d}_{\nabla_x E_{\rm PCN}}
    \;+\; \underbrace{\lambda d_\beta}_{\rm tether}
    \;+\; \underbrace{\beta\big(d_\beta - \tfrac{\gamma}{\lambda}u\big)}_{\rm nudge}
    \;=\; 0
    \quad\Longrightarrow\quad
    d_\beta = \frac{\lambda d + \beta\gamma u/\lambda}{\lambda + \beta} ,
\end{equation}
and subtracting $d$ gives
\begin{equation}
    d_\beta - d = \frac{\beta\big(\gamma u/\lambda - d\big)}{\lambda + \beta} ,
\end{equation}
into which $d = \gamma v^*/\lambda$ from \eqref{eq:velocity-readout} and
$g = v^* - u$ substitute to give
\begin{equation}\label{eq:quad-displacement}
    x^{*\beta} - x^* = d_\beta - d = -\frac{\beta\gamma\, g}{\lambda(\lambda + \beta)}.
\end{equation}
This is nonlinear in $\beta$ through
the $\lambda + \beta$ in the denominator, and linear only while
$\beta \ll \lambda$. Replacing $\beta$ by $-\beta$ gives a displacement of different magnitude,
$\lambda - \beta$ in place of $\lambda + \beta$, so the $\pm\beta$ equilibria
are not symmetric about $x^*$, and the
centred estimator \eqref{eq:ep-update} inherits a bias the next paragraphs put
at $O((\beta/\lambda)^2)$.

\emph{Linearised cost.} This targets the same gradient as the quadratic cost:
$g$ is a constant during the nudged phases, so $\nabla_\theta
C_{\rm lin}(x^*) = g^{\top}\nabla_\theta v^*$, which is the chain rule for
$\nabla_\theta \tfrac12\|v^* - u\|^2$. 

Here
\begin{equation}
    \beta' C_{\rm lin}(x) = \beta\Big(\tfrac{\gamma}{\lambda}\Big)^2
        \tfrac{\lambda}{\gamma}\, g^{\top}(x - x_t)
        = \beta\,\tfrac{\gamma}{\lambda}\, g^{\top}(x - x_t) ,
    \qquad
    \nabla_x\,\beta' C_{\rm lin} = \beta\,\tfrac{\gamma}{\lambda}\, g ,
\end{equation}
a constant force independent of $x$, so stationarity reads
\begin{equation}
    \underbrace{-\lambda d}_{\nabla_x E_{\rm PCN}}
    \;+\; \underbrace{\lambda d_\beta}_{\rm tether}
    \;+\; \underbrace{\beta\,\tfrac{\gamma}{\lambda}\, g}_{\rm nudge}
    \;=\; 0
    \quad\Longrightarrow\quad
    d_\beta = d - \frac{\beta\gamma\, g}{\lambda^2} ,
\end{equation}
and subtracting $d$ gives
\begin{equation}\label{eq:lin-displacement}
    x^{*\beta} - x^* = d_\beta - d = -\frac{\beta\gamma\, g}{\lambda^2}.
\end{equation}
There is no $\beta$ in the denominator: the displacement is linear in $\beta$,
so the $\pm\beta$ equilibria are symmetric about $x^*$ up to the
$O(\gamma/\lambda)$ accuracy of holding $\nabla_x E_{\rm PCN}$ fixed. The other $\beta$-dependent error that remains comes
from $\partial_\theta E_{\rm PCN}$ varying nonlinearly along the straight path
between the two equilibria, which the next paragraphs bound.

\paragraph{The centred estimate.} Write $\Phi(x) = \partial_\theta
E_{\rm PCN}(x, h^*(x))$ for the local gradient at the hidden equilibrium for a
given visible state, and $G(\beta) = \Phi(x^{*\beta})$. Expanding
$G(\pm\beta)$ about $\beta = 0$ and subtracting cancels the even orders,
\begin{equation}
    \frac{G(\beta) - G(-\beta)}{2\beta}
    = G'(0) + \frac{\beta^2}{6}G'''(0) + O(\beta^4) ,
\end{equation}
so the estimate in \eqref{eq:ep-update} returns the wanted $G'(0)$ with a
relative error $\beta^2|G'''(0)|/6|G'(0)|$, componentwise. What differs between
the two costs is where that third derivative comes from.

\paragraph{Nonlinearity in $\beta$ under the quadratic cost.} From
\eqref{eq:quad-displacement} the nudged state moves along the fixed direction
$-g$ at a rate nonlinear in $\beta$,
\begin{equation}
    x^{*\beta} = x^* + f(\beta)\, w_0 , \qquad
    w_0 = -\frac{\gamma g}{\lambda} , \qquad
    f(\beta) = \frac{\beta}{\lambda + \beta} ,
\end{equation}
and by the chain rule
\begin{align}
    G'(0)   &= f'(0)\,\nabla_x\Phi\cdot w_0 , \\
    G'''(0) &= f'''(0)\,\nabla_x\Phi\cdot w_0
             + 3f'(0)f''(0)\,\nabla_x^2\Phi[w_0, w_0]
             + f'(0)^3\,\nabla_x^3\Phi[w_0, w_0, w_0] .
\end{align}
Three things can bias the estimate: the nonlinearity of $f$, giving the first
term of $G'''(0)$; the curvature of $\Phi$, giving the other two, which carry
its second and third derivatives; and the approximation of the previous section
that held $\nabla_x E_{\rm PCN}$ fixed, a relative $O(\gamma/\lambda)$ error in
the equilibria themselves. The last does not grow with $\beta$ and so cannot
set a ceiling on it. Of the other two, the first term shares the factor
$\nabla_x\Phi\cdot w_0$ with $G'(0)$, so
\begin{equation}
    \frac{\beta^2}{6}\,\frac{f'''(0)\,\nabla_x\Phi\cdot w_0}{f'(0)\,\nabla_x\Phi\cdot w_0}
    = \frac{\beta^2}{6}\,\frac{6/\lambda^3}{1/\lambda}
    = \Big(\frac{\beta}{\lambda}\Big)^{\!2} ,
\end{equation}
a relative error present for any $\Phi$ and at any $\gamma$. The two $\Phi$-curvature terms of $G'''(0)$ are smaller than
the first by $\|w_0\|$ and $\|w_0\|^2$ respectively: each extra $w_0$ costs a
factor $\|w_0\| = \gamma|g|/\lambda$, while the $O(\gamma)$ amplitudes of the
$\nabla_x^n\Phi$ themselves are common to all three terms and cancel (next
paragraph). Those ratios are independent of $\beta$, and at the small $\gamma$
the correspondence requires they are tiny --- $\|w_0\| \approx 4\times10^{-5}$
at the settings of Section~\ref{subsec:equivalence-at-training} --- so under the quadratic cost it is $f$'s nonlinearity that limits $\beta$.
This is the asymmetry of the $\pm\beta$ equilibria noted above, now quantified:
requiring $(\beta/\lambda)^2 \ll 1$ is the constraint $\beta \ll \lambda$ of
\eqref{eq:lin-constraint}.

\paragraph{Nonlinearity in $\beta$ under the linearised cost.} From
\eqref{eq:lin-displacement} the path is $x^{*\beta} = x^* + f(\beta)\,w_0$ with
the same direction $w_0 = -\gamma g/\lambda$, but now $f(\beta) = \beta/\lambda$
exactly. Every derivative of $f$ beyond the first vanishes, so the first two
terms of $G'''(0)$ go with them and
\begin{equation}
    G'(0) = \frac{1}{\lambda}\,\nabla_x\Phi\cdot w_0 , \qquad
    G'''(0) = \frac{1}{\lambda^3}\,\nabla_x^3\Phi[w_0, w_0, w_0] .
\end{equation}
Only $\Phi$'s own curvature is left. Its size follows from the same argument as
before: every error at the hidden equilibrium is $O(\gamma)$
(App.~\ref{app:derivation}, Step 2) and $\Phi$ is bilinear in errors and
activations, so $\Phi$ has amplitude $O(\gamma)$, while the distance in $x$ over
which it changes appreciably is set by the $O(1)$ weights and nonlinearities, so
$\|\nabla_x^n\Phi\| = O(\gamma)$ for every $n$. Those $O(\gamma)$ amplitudes
cancel in the ratio, leaving two extra factors of $w_0$ and two of $\lambda$:
\begin{equation}
    \frac{\beta^2 |G'''(0)|}{6|G'(0)|}
    = O\!\left(\frac{\beta^2\|w_0\|^2}{\lambda^2}\right)
    = O\!\left(\Big(\frac{\beta\gamma|g|}{\lambda^2}\Big)^{\!2}\right)
    = O\big(|x^{*\beta} - x^*|^2\big) :
\end{equation}
the estimate is accurate while the nudge displacement is small on the scale
over which $\Phi$ varies. Unlike the quadratic cost's bias this is a condition
on the displacement, which smaller $\gamma$ relaxes rather than leaves
untouched, and requiring it to be small is the constraint $\beta\gamma|g| \ll \lambda^2$
of \eqref{eq:lin-constraint}.

\paragraph{Consistency with measurement.} Both displacement formulas reproduce
the diagnostics logged during training to three significant figures. At the
operating point ($\gamma = 10^{-4}$, $\lambda = 1$, $\beta = 30$), with the
logged per-element RMS values $|v^*| = 1.03$ and $|g| = 0.39$,
\eqref{eq:velocity-readout} predicts a free-phase extension
$|x^* - x_t| = 1.03\times10^{-4}$ and \eqref{eq:lin-displacement} a nudge
displacement of $1.16\times10^{-3}$; the logged values are $1.03\times10^{-4}$
and $1.16\times10^{-3}$. The two constraints then place the ceilings
four orders of magnitude apart: $\beta \ll \lambda = 1$ for the quadratic
cost, against $\beta \ll \lambda^2/(\gamma|g|) = 2.6\times10^4$ for the
linearised one. Section~\ref{subsec:equivalence-at-training} measures the
quadratic nudge failing above $\beta = 1$ and the linear nudge degrading only
gently towards $10^4$. 

\paragraph{Scope.} The linearisation does not depend on the form of
$E_{\rm PCN}$, so it applies to any use of GradEP with a quadratic loss. It is
also simpler to realise physically: a constant force on the visible nodes,
rather than an added quadratic potential whose restoring force must track the
displacement.

\section{Predictive coding on a DAG}\label{app:dag-pcn}

\subsection{From chains to DAGs}

Predictive coding networks are usually written on a \emph{chain}, with node $k$
predicted from node $k-1$ alone:
\begin{equation}\label{eq:chain-pcn}
    E_{\text{chain}} = \sum_{k=1}^{N} \tfrac{1}{2}\big\| f_k(h_{k-1}) - h_k \big\|^2 ,
\end{equation}
with $h_0$ the clamped input. The UNet we adopt is not a chain: its decoder
blocks consume a concatenation of their predecessor and a topologically distant
encoder state, so a node may have several parents, and an encoder node several
children. Predictive coding has been formulated on arbitrary computation graphs~\citep{millidgePredictiveCodingApproximates2022}, and we use that construction unchanged:
nodes $h_1, \dots, h_N$ are taken in topological order, each with a parent list
$\mathrm{pa}(k)$, and
\begin{equation}\label{eq:dag-pcn}
    E = \sum_{k=1}^{N} \tfrac{1}{2}\big\| f_k\big(h_{\mathrm{pa}(k)}\big) - h_k \big\|^2
        \; + \; \gamma\alpha\, o ,
\end{equation}
with $o = h_N$ the scalar output node and $\gamma\alpha$ the injection of
Section~\ref{subsec:correspondence}. The prediction functions are the
unmodified architecture blocks; where a node has several parents they are
concatenated in the order of the feedforward pass, so $f_k$ evaluates the same
function of the same inputs. 

\subsection{Relaxation dynamics}

Writing $e_k = f_k(h_{\mathrm{pa}(k)}) - h_k$ and
$J_{c,j} = \partial f_c / \partial h_j$, each state evolves under
\begin{equation}\label{eq:dag-dynamics}
    \frac{\partial E}{\partial h_j}
    = -\,e_j
      \; + \sum_{c \,\in\, \mathrm{ch}(j)} J_{c,j}^{\top}\, e_c
      \; + \; \gamma\alpha\, \delta_{jN} ,
\end{equation}
the node's own error term, the error term of every child that consumes it, and
the output injection. In a chain $\mathrm{ch}(j) = \{j+1\}$; in the DAG a node
with fan-out is driven by all its children at once, and stationarity is the
multivariable chain rule (App.~\ref{app:derivation}). Every term lives on an
edge of the graph, so locality is preserved.

\subsection{Node granularity}\label{subsec:granularity}

How a network is partitioned into nodes is a free choice. A node can be an
architecture \emph{block} --- one residual block, one attention block, or the
ViT head etc --- in which case only the block's output relaxes and its internal
activations are computed feedforward inside $f_k$; or it can be a single weight
matrix and its activation, in which case every hidden activation relaxes. The
correspondence of Section~\ref{subsec:correspondence} holds at any partition.
What the choice sets is how the backward pass is divided between relaxation and
circuitry: finer partitions relax more states and need less backward circuitry,
which suits hardware; coarser ones relax fewer states and need more backpropagation, which
suits simulation. For a parameter inside node $k$,
\begin{equation}
    \frac{\partial E}{\partial \theta}
    = e_k^\top \frac{\partial f_k}{\partial \theta} .
\end{equation}
At one node per weight matrix, $f_c(h_j) = \sigma(W h_j)$, the update is an
outer product of quantities at the two ends of the weight,
$\partial E/\partial W = (\sigma' \odot e_c)\,h_j^\top$, and the backward
drive of \eqref{eq:dag-dynamics} is $J_{c,j}^\top e_c = W^\top(\sigma' \odot
e_c)$: the child's error gated by the activation derivative and passed back
through the transposed weights. For a multi-layer block, the same backward drive is a
vector--Jacobian product through every layer inside it, which must be provided
by additional circuitry (Section~\ref{sec:discussion}). At the coarsest extreme
the whole network is one node, the update reduces to
$\gamma\alpha\,\nabla_\theta f_1$ exactly, and the energy is a wrapper around a
full-network backpropagation computation.\footnote{With a single node, $h_1 = o$ and
$E = \tfrac12\|f_1(x) - h_1\|^2 + \gamma\alpha\, h_1$. Stationarity in $h_1$
gives $e_1 = \gamma\alpha$ exactly, and $\partial E/\partial\theta = e_1
\nabla_\theta f_1 = \gamma\alpha\, \nabla_\theta f_1$.}

Accuracy runs the other way. The corrections to the correspondence come from
interior nodes, whose equilibria sit $O(\gamma)$ off the forward pass
(App.~\ref{app:derivation}), so a finer partition has more correction terms.
They vanish as $\gamma \to 0$ at every partition, but granularity sets how small
$\gamma$ must be for a given accuracy.

\paragraph{Measured at two granularities.} We compare block granularity ($27$
nodes) with weight-matrix granularity ($74$ nodes) on the CIFAR-10 replication
checkpoint; the two graphs compute an identical forward pass, and all
measurements here are in float64. With a generous relaxation budget
($K_h = 3$, $T_{\rm free} = 60$, $T_{\rm nudge} = 30$),
$\cos(\Delta\theta_{\rm EP}, \Delta\theta_{\rm BP})$ is $1.0000$ to four
decimals at both granularities and at every $\gamma$ from $3\times10^{-7}$ to
$10^{-4}$. At the budget of our EP training run (Table~\ref{tab:hyperparams}),
both give $0.9999$ at $\gamma = 10^{-4}$ and $1.0000$ at
$\gamma = 3\times10^{-5}$ and $10^{-5}$, with update norms within $0.2\%$ of
each other; both are about $2\%$ shorter than the backpropagation gradient, a
consequence of the budget rather than the partition. The finer partition does
raise the correspondence error, as the previous paragraph predicts: the
relative velocity deviation is linear in $\gamma$ at both, with a coefficient
$13.7$ times larger at the finer one ($2.04\times10^{-4}$ against
$1.49\times10^{-5}$ at $\gamma = 10^{-4}$). That places the finer partition's
bias at the TF32 floor of Figure~\ref{fig:gamma-floors} rather than below it,
so in that arithmetic its
operating $\gamma$ would need to be smaller. At inference both partitions converge in a single sweep in error coordinates
($K_h = 1$ at $\mathrm{d}t = 1$, App.~\ref{app:error-param}), which costs
$7$--$8\%$ more at the finer one on CPU. These are single-checkpoint
measurements; we did not train at weight-matrix granularity, so we cannot say
whether its training dynamics differ over many steps.

\subsection{The adopted graph}

For the runs of the main text we use block granularity, so that whole modules
of the backbone serve as $f_k$ unmodified. The graph has $N = 27$ nodes: twelve
encoder blocks chained from the input, the middle block, twelve decoder blocks
each taking $[\text{previous}, \text{skip}]$ as parents with the skip popped
from the encoder stack in the architecture's own order, the output projection,
and the ViT head producing $o$. The time-conditioning pathway inherited from
the diffusion backbone is not a node: $V(x)$ is time-independent, so it
receives a fixed constant, which each block sees as a trained per-channel
bias.\footnote{The feedforward reference does the same; injecting the identical
constant is what preserves exact agreement with its forward pass. The pathway
carries no state and does not relax.}

\section{Error-parameterised relaxation}\label{app:error-param}

Rather than relaxing the node states $h_k$ directly, we follow
\citet{goemaereEPCOvercomingExponential2025} and relax the prediction errors
$e_k = f_k(h_{\mathrm{pa}(k)}) - h_k$, recovering states by the topological
scan $h_k = f_k(h_{\mathrm{pa}(k)}) - e_k$. This is an invertible change of
variables on the same energy, so the two parameterisations share their
stationary points. Their curvature, however, is very different, and at the
step sizes of a practical simulation only one of them converges in a usable
number of sweeps. The error parameterisation is a simulation device rather than dynamics a
substrate would run: $\partial E/\partial e$ is a reverse-mode derivative
through the scan, and a single sweep from $e = 0$ at $\mathrm{d}t = 1$ is one
backward pass \citep[Thm.~C.9]{goemaereEPCOvercomingExponential2025}. What it
locates, however, is the fixed point of $E_{\rm PCN}$, which does not depend on
how it is found: the local state dynamics \eqref{eq:dag-dynamics} converge to
the same point (below), and every result in this paper is a property of that
point.

In the $h$-parameterisation each state enters both its own error term and the
error term of every child that consumes it, so $\partial^2 E/\partial h^2$ has
diagonal blocks $I + \sum_{c} J_{c,j}^\top J_{c,j}$ (plus second-derivative
terms in $f_c$, suppressed by $e_c = O(\gamma)$) and off-diagonal blocks
$-J_{c,j}^\top$ for each parent--child pair: its curvature is set by the squared
Jacobians of the architecture. In the $e$-parameterisation the errors are independent coordinates, so
$\sum_k \tfrac12 \|e_k\|^2$ has Hessian exactly $I$. The network's layers now
enter only through the injection: $h_N$ is recovered from the errors by the
forward scan, so it is a nonlinear function of every $e_k$, and the injection
$\gamma\alpha\, h_N$ contributes that function's curvature, scaled by
$\gamma$:
\begin{equation}
   \frac{\partial^2 E}{\partial e^2} = I + \gamma\, M ,
    \qquad M = \alpha \nabla_e^2\, h_N ,
\end{equation}
with $M$ independent of $\gamma$.

\paragraph{Measured curvature.} On the CIFAR-10 replication checkpoint at the
operating $\gamma = 10^{-4}$, power iteration on Hessian--vector products gives
a largest Hessian eigenvalue $\mu_{\max}$ of $1.0001$ in error coordinates and
at least $8\times10^{4}$ in state coordinates. Gradient descent at step
$\mathrm{d}t$ is stable only for $\mathrm{d}t < 2/\mu_{\max}$: below $2$ in
error coordinates, which admits the deadbeat step $\mathrm{d}t = 1$, and below
$2.4\times10^{-5}$ in state coordinates.

\paragraph{The state parameterisation is impractically slow.} At
$\mathrm{d}t = 1$ the state relaxation diverges within a few sweeps, in float64
as in float32, so the failure is a dynamical instability rather than a
precision floor. Below the stability bound it does converge, monotonically and
to the same fixed point as the error relaxation, as a shared stationary point
requires. But it converges at a measured rate implying a condition number of about $2\times10^{7}$: at $\mathrm{d}t = 2\times10^{-5}$, reaching equilibrium
takes of the order of $10^{8}$ sweeps, against one in error coordinates. This
cost is a property of the explicit discretisation rather than of the dynamics:
a continuous system settles at the rate of its slowest mode, not its fastest,
and \citet[App.~B.5]{goemaereEPCOvercomingExponential2025} make the same point, that
the decay which makes discrete state updates slow is an artefact of time
discretisation.

\paragraph{Consequences.} The error parameterisation is therefore not just a
convenience but what makes the relaxation of this network practical. Three
further properties follow from its near-identity Hessian, each measured
elsewhere in the paper. The relaxation reaches equilibrium in a single deadbeat
sweep from feedforward initialisation
(App.~\ref{app:correspondence-measurement}). The implicit-gradient solve of
Section~\ref{subsec:training} inverts a matrix already close to the identity,
so one conjugate-gradient step is as accurate as fifty
(App.~\ref{app:training-fidelity}). Finally, float32 suffices: at the operating
$\gamma$ the velocity's deviation from autograd's sits below the arithmetic's
own floor (Section~\ref{subsec:equivalence-at-inference}). Part of the reason is that the error parameterisation never forms an error as
the difference of two nearly equal states, which in state coordinates costs
significant digits of precision in every error.

\section{Measuring the velocity correspondence}
\label{app:correspondence-measurement}

\paragraph{Measurement.} The PCN relaxation velocities $v_{\rm PCN}$ and the
FFN autograd velocities $v_{\rm FFN}$ are both evaluated in the stated
arithmetic, and each saturates at its own floor. The highest is
$\sim 2\times10^{-4}$, for float32 tensors whose multiplications run in
truncated TF32 precision on the tensor cores --- the default A100 math mode,
under which all our CIFAR-10 training and FID runs were performed. Forcing
those multiplications to full float32 lowers it to $\sim 4\times10^{-7}$, and
float64 does not saturate anywhere in the measured range, though it requires
patching the adopted architecture's GroupNorm wrapper, which otherwise casts
to float32 unconditionally and would impose a spurious float32 floor.

\begin{figure}[t]
\centering
\begin{subfigure}[t]{0.48\linewidth}
\centering
\includegraphics[width=\linewidth]{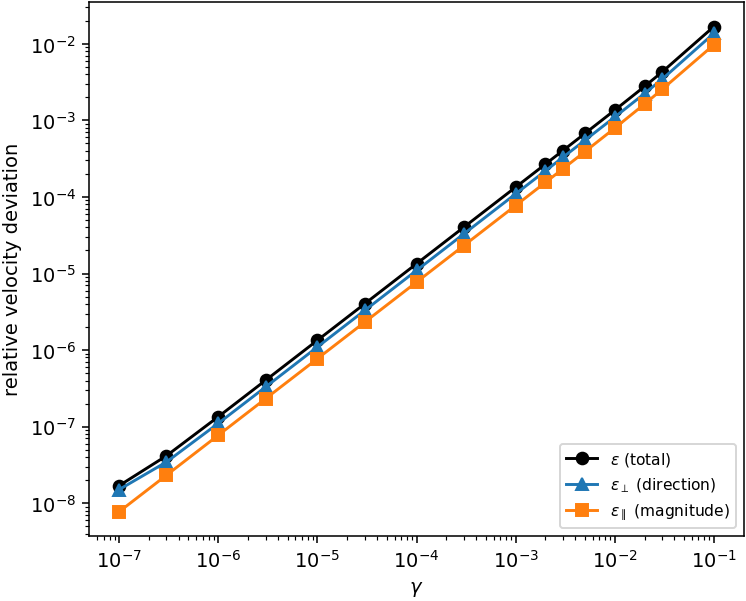}
\caption{Direction/magnitude decomposition.}
\label{fig:direction-magnitude}
\end{subfigure}
\hfill
\begin{subfigure}[t]{0.48\linewidth}
\centering
\includegraphics[width=\linewidth]{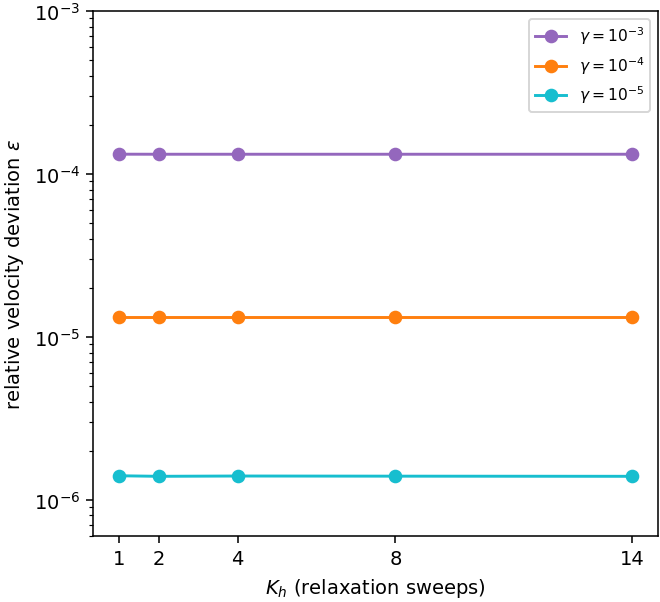}
\caption{Budget independence.}
\label{fig:kh-flatness}
\end{subfigure}
\caption{Decomposition and budget independence of the velocity deviation,
measured as in Figure~\ref{fig:gamma-floors}. \textbf{(a)} The relative
velocity deviation
$\epsilon = \|v_{\rm PCN} - v_{\rm FFN}\| / \|v_{\rm FFN}\|$ resolved into
components parallel (magnitude error) and perpendicular (direction error) to
$v_{\text{FFN}}$. Both scale as $O(\gamma)$, with direction the larger,
$\epsilon_\perp^2$ carrying roughly two-thirds of $\epsilon^2$ (the curves combine in quadrature).
\textbf{(b)} At $\mathrm{d}t = 1$ the deviation is independent of the number of
relaxation sweeps $K_h$ at every $\gamma$ tested: a single sweep already
reaches the equilibrium.}
\label{fig:correspondence-details}
\end{figure}

\paragraph{Direction and magnitude.} Consider a single interpolant, and let
$\phi$ be the angle between $v_{\text{PCN}}$ and $v_{\text{FFN}}$. Resolving $v_{\text{PCN}}$ along and
perpendicular to $v_{\text{FFN}}$ gives components
$\|v_{\text{PCN}}\|\cos\phi$ and $\|v_{\text{PCN}}\|\sin\phi$, so the
deviation $\Delta v = v_{\text{PCN}} - v_{\text{FFN}}$ has
\begin{equation}
    \|\Delta v_\parallel\| = \big|\, \|v_{\text{PCN}}\|\cos\phi - \|v_{\text{FFN}}\| \,\big| ,
    \qquad
    \|\Delta v_\perp\| = \|v_{\text{PCN}}\|\sin\phi .
\end{equation}
For $\phi \ll 1$ these reduce to $\big|\,\|v_{\text{PCN}}\| - \|v_{\text{FFN}}\|\,\big|$
and $\|v_{\text{PCN}}\|\,\phi$: the parallel component is an error in the
velocity's magnitude, the perpendicular component an error in its direction. Normalising both by $\|v_{\text{FFN}}\|$ gives the relative errors
$\epsilon_\parallel \approx \big|\,\|v_{\text{PCN}}\|/\|v_{\text{FFN}}\| - 1\big|$
and $\epsilon_\perp \approx \phi$, the second using
$\|v_{\text{PCN}}\|/\|v_{\text{FFN}}\| = 1 + O(\gamma)$. Since the components
are orthogonal,
\begin{equation}
    \epsilon^2 = \epsilon_\parallel^2 + \epsilon_\perp^2 ,
\end{equation}
where $\epsilon$ is the total relative deviation of
Figure~\ref{fig:gamma-floors}. Figure~\ref{fig:direction-magnitude} plots all
three: both components are $O(\gamma)$, with direction carrying roughly
two-thirds of $\epsilon^2$ across the range. At $\gamma = 10^{-4}$,
$\phi \approx 1.1\times10^{-5}$\,rad, i.e.\
$\cos(v_{\text{FFN}}, v_{\text{PCN}}) = 1$ to nine decimal places.

\paragraph{Deadbeat solver.} At inference the errors $e_k$ are relaxed for
$K_h$ sweeps at $\mathrm{d}t = 1$, and Figure~\ref{fig:kh-flatness} shows the
deviation is independent of $K_h$: values in $\{1, 2, 4, 8, 14\}$ agree to within $1\%$. This follows from three choices acting together.
The error parameterisation makes the relaxation Hessian a perturbation of the
identity, $\partial^2 E/\partial e^2 = I + \gamma M$
(Appendix~\ref{app:error-param}); gradient descent with $\mathrm{d}t = 1$ on
identity curvature is deadbeat, the iteration map $I - \mathrm{d}t\,H =
-\gamma M$ annihilating the error in a single sweep up to $O(\gamma)$; and feedforward initialisation\footnote{Only the single sweep ($K_h = 1$) depends on this: from
unit-Gaussian random initial hidden states --- $O(1)$ initial errors, against
$O(\gamma)$ at equilibrium --- $K_h = 1$ fails to converge, but $K_h \geq 5$
reproduces the feedforward-initialised velocity exactly, unchanged out to
$K_h = 60$. The equilibrium is therefore a genuine attractor of the relaxation,
not an artifact of the feedforward initialisation.} starts the
system at $e_k = 0$, the exact minimum of the $\gamma = 0$ energy, so only the
$O(\gamma)$ displacement the injection causes remains. The first sweep
therefore lands within $O(\gamma^2)$ of the equilibrium errors. We read the
velocity as $\nabla_x E_{\rm PCN}$ at fixed $e$. With the errors held fixed,
only the injection term depends on $x$, so
$v_{\rm PCN} = -\gamma^{-1}\nabla_x E_{\rm PCN} = -\alpha\,\nabla_x h_N(x, e)$:
the $\gamma$ cancels, and $e$ enters as the argument of an $O(1)$ function
rather than being divided by $\gamma$. An $O(\gamma^2)$ error in $e$ therefore
moves the velocity by only $O(\gamma^2)$, against the $O(\gamma)$ bias, so the
residual deviation in Figure~\ref{fig:gamma-floors} is that bias rather than
under-relaxation, as the flatness in $K_h$ confirms. At equilibrium this
readout equals the $-\gamma^{-1}J_1^\top e_1^*$ of
Section~\ref{subsec:correspondence} by the envelope identity; away from
equilibrium the two differ. The flatness from $K_h = 1$ is specific to $\mathrm{d}t = 1$; smaller steps
contract the error by a constant factor each sweep, reaching the same fixed
point over a few sweeps. Convergence of the $\mathrm{d}t = 1$ iteration requires the
eigenvalues of $\gamma M$ to remain below unity, a bound never approached in the
measured range. Note though that this is a constraint of the explicit Euler
discretisation rather than of the model: continuous-time hardware has no step
size, and relaxes by true gradient flow at any $\gamma$. All inference
measurements in the main text therefore use $K_h = 1$ at $\mathrm{d}t = 1$
from feedforward initialisation.

\begin{figure}[t]
\centering
\includegraphics[width=0.75\linewidth]{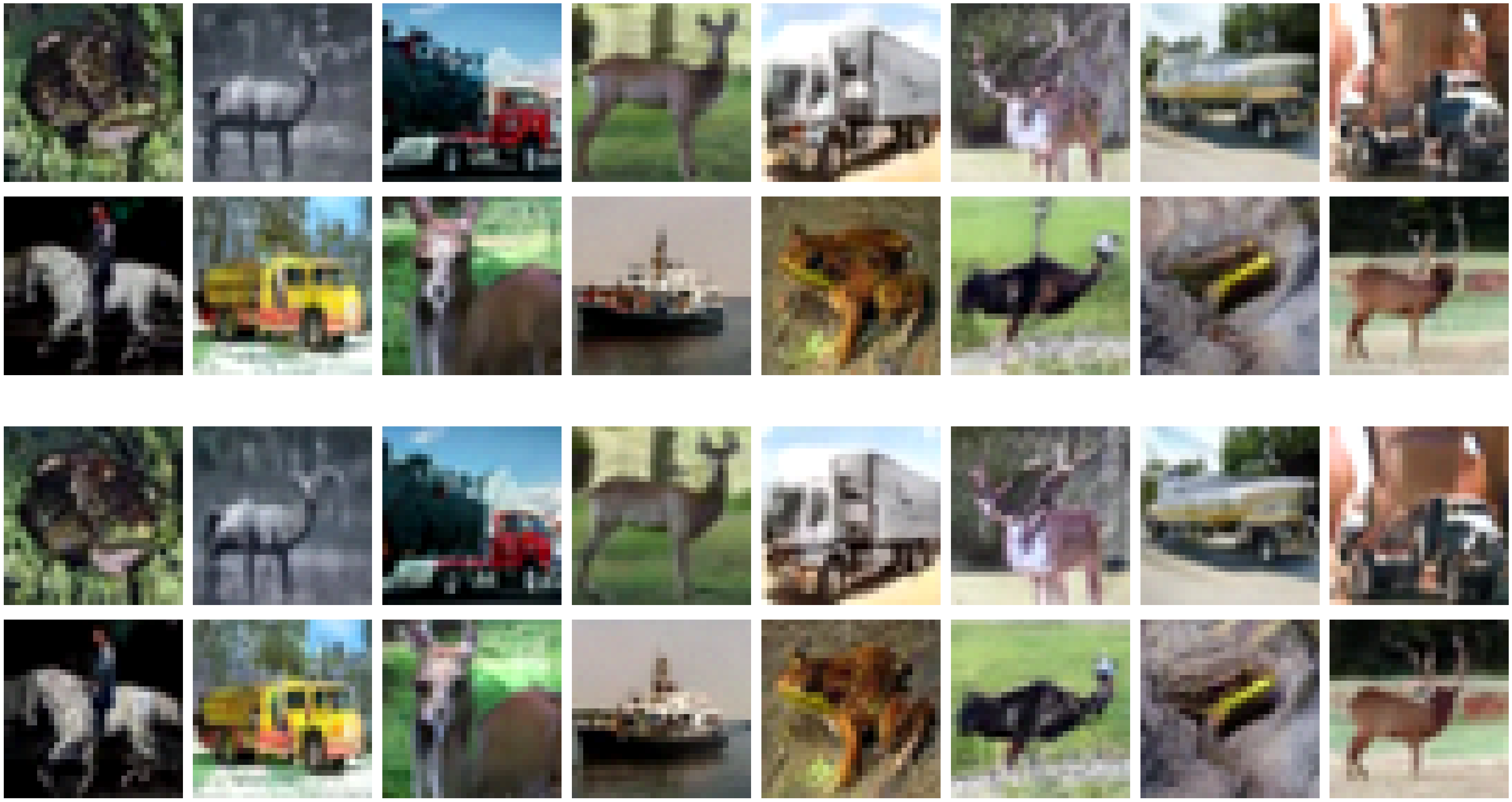}
\caption{The same sixteen noise realisations sampled by feedforward autograd
velocities (top) and by PCN relaxation velocities (bottom), on our CIFAR-10
phase-2 replication checkpoint at $\gamma = 10^{-4}$ and $\tau_s = 3.25$.}
\label{fig:cifar-ffn-vs-pcn}
\end{figure}

\paragraph{Samples.} Figure~\ref{fig:cifar-ffn-vs-pcn} shows the same sixteen noise
realisations sampled by each of the $v_{\rm FFN}$ and $v_{\rm PCN}$ velocity sources; the pairs are visually
indistinguishable.

\section{Transfer to ImageNet-32}\label{app:imagenet-transfer}

The ImageNet-32 checkpoints use the same architecture and output scale as the
CIFAR-10 model, so the PCN construction and the $\gamma$ convention carry over
unmodified; only the trained potential $V(x)$ differs. The ImageNet recipe
additionally applies a saturating output clamp, sampling with
$-\nabla_x\,c\tanh(V/c)$ for $c = 10^4$. Since the output node carries the
potential pre-scale ($V = \alpha o$), we absorb the clamp into the top
prediction function as $f_N' = c\tanh(\alpha f_N/c)/\alpha$, so that
$\alpha o' = c\tanh(V/c)$ and the injection term $\gamma\alpha o'$ acts on exactly
the quantity the feedforward sampler differentiates. The correspondence then
applies unchanged, with no additional term.

\begin{figure}[t]
\centering
\includegraphics[width=0.75\linewidth]{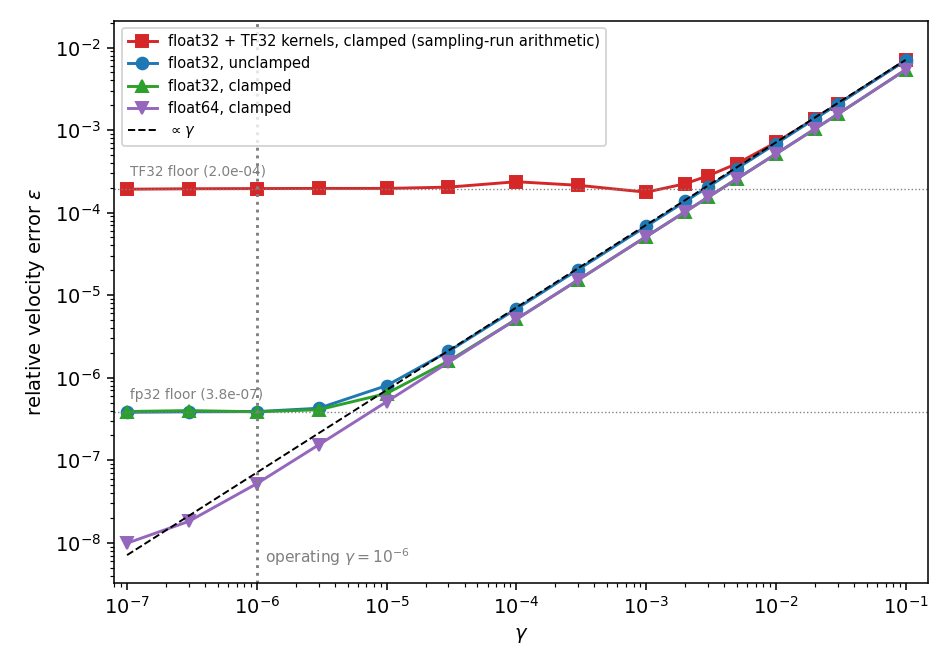}
\caption{Relative velocity deviation $\epsilon$ on the authors' phase-1
ImageNet-32 checkpoint, measured as in Figure~\ref{fig:gamma-floors}. The
$O(\gamma)$ law and both arithmetic floors reproduce their CIFAR-10 values,
and the clamped potential, with the clamp incorporated into the top
prediction function, follows the same law and shares the same floor. At the operating $\gamma = 10^{-6}$
(dotted) the deviation lies far below the floor of the sampling-run
arithmetic.}
\label{fig:imagenet-transfer}
\end{figure}

Figure~\ref{fig:imagenet-transfer} repeats the sweep of
Figure~\ref{fig:gamma-floors} on these checkpoints. The $O(\gamma)$ law holds
over the same range, up to $\gamma = 0.1$; the float32 and TF32 floors sit at
$3.8\times10^{-7}$ and $2.0\times10^{-4}$, within ten per cent of their
CIFAR-10 values, and the three arithmetics coincide above $\gamma \sim
10^{-2}$ as before. The clamped and unclamped potentials follow the same law.
At the operating point $\gamma = 10^{-6}$ the $O(\gamma)$ bias is
$\sim 5\times10^{-8}$, some four orders of magnitude below the TF32 floor of
the sampling runs.

\section{Training fidelity: budgets and nudge strength}
\label{app:training-fidelity}

\begin{figure}[t]
\centering
\includegraphics[width=\linewidth]{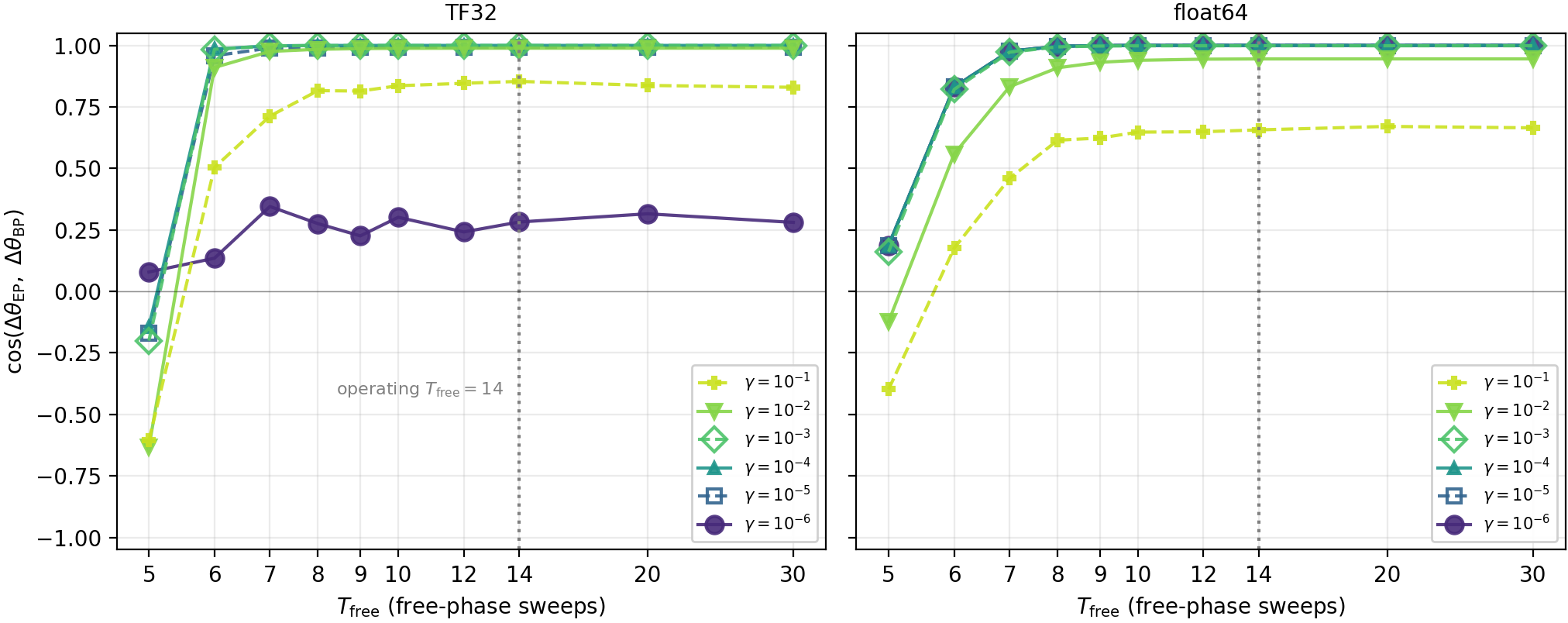}
\caption{$\cos(\Delta\theta_{\rm EP}, \Delta\theta_{\rm BP})$ against the
free-phase relaxation budget $T_{\rm free}$, at six values of $\gamma$, in
TF32 (left) and float64 (right). CIFAR-10 replication checkpoint,
$512$ flow-matching interpolants, $T_{\rm nudge} = 6$, $K_h = 1$,
$\beta = 30$. Every $\gamma$ that reaches full fidelity does so at the same
budget; the plateau heights, by contrast, are specific to this $\beta$
(Table~\ref{tab:gamma-beta}).}
\label{fig:train-budget}
\end{figure}

\paragraph{The free-phase threshold.} Figure~\ref{fig:train-budget} plots
$\cos(\Delta\theta_{\rm EP}, \Delta\theta_{\rm BP})$ against the free-phase
budget $T_{\rm free}$. Where the free phase has not converged the update
correlates poorly with backpropagation's, and the recovery is
abrupt. The level each curve then plateaus at is a separate matter, set by
$\gamma$ and $\beta$ together with the arithmetic
(Table~\ref{tab:gamma-beta}); here we are concerned only with where the
transition happens. Across the TF32 grid we measured --- $\gamma$ from
$10^{-6}$ to $10^{-1}$ at $\beta = 30$, and $\beta \in \{3, 100\}$ at
$\gamma = 10^{-5}$, $10^{-4}$ and $10^{-2}$ --- every cell that reaches full
fidelity lies between $-0.20$ and $-0.10$ at $T_{\rm free} = 5$ where
measured, is at least $0.958$ at $T_{\rm free} = 6$, and is above $0.99$ by
$T_{\rm free} = 9$. Neither $\gamma$ nor $\beta$ moves the transition within
this range. At $\gamma = 10^{-5}$ the cosine plateaus at about $0.993$; below
that the plateau falls to $0.978$ at $\gamma=5\times10^{-6}$ and never exceeds
$0.35$ at $\gamma=10^{-6}$, where the update is arithmetic-limited and there is no
transition to full fidelity to locate. We
therefore use $T_{\rm free} = 14$ throughout, well past the transition, and
$T_{\rm nudge} = 6$, above what the next paragraphs show is needed.

\begin{table}[t]
\centering
\caption{Parameter-update fidelity
$\cos(\Delta\theta_{\rm EP},\Delta\theta_{\rm BP})$ over nudge strength
$\beta$ (rows) and injection strength $\gamma$ (columns), varied about the
settings of our EP phase-1 training run (Table~\ref{tab:hyperparams}), in
TF32 (the arithmetic of our training runs), on the
CIFAR-10 replication checkpoint at $512$ flow-matching interpolants. Shaded:
$\cos < 0.99$. Every failure in the grid belongs to the EP estimator, not the
correspondence: the implicit solve, which uses no nudge, is exact to four
decimal places up to $\gamma = 10^{-2}$.}
\label{tab:gamma-beta}
\begin{tabular}{r cccccc}
\toprule
$\beta$ & $10^{-5}$ & $3\times10^{-5}$ & $10^{-4}$ & $10^{-3}$ & $10^{-2}$ & $10^{-1}$ \\ \midrule
0.3 & \cellcolor{gray!15}0.0803 & \cellcolor{gray!15}0.2353 & \cellcolor{gray!15}0.7701 & 0.9933 & \cellcolor{gray!15}0.9896 & \cellcolor{gray!15}0.3391 \\
1 & \cellcolor{gray!15}0.3437 & \cellcolor{gray!15}0.7149 & \cellcolor{gray!15}0.9536 & 0.9988 & 0.9952 & \cellcolor{gray!15}0.8840 \\
3 & \cellcolor{gray!15}0.6657 & \cellcolor{gray!15}0.9459 & 0.9924 & 0.9996 & 0.9960 & \cellcolor{gray!15}0.3935 \\
10 & \cellcolor{gray!15}0.9489 & 0.9920 & 0.9991 & 0.9996 & 0.9949 & \cellcolor{gray!15}0.6788 \\
30 & 0.9937 & 0.9989 & 0.9998 & 0.9990 & \cellcolor{gray!15}0.9836 & \cellcolor{gray!15}0.3902 \\
100 & 0.9990 & 0.9998 & 0.9997 & 0.9969 & \cellcolor{gray!15}0.9722 & \cellcolor{gray!15}0.2807 \\
\midrule
\multicolumn{7}{l}{\itshape controls} \\
$\Delta\theta_{\rm IFT}$, TF32 & 1.0000 & 1.0000 & 1.0000 & 1.0000 & 1.0000 & 0.9976 \\
$\Delta\theta_{\rm EP}$, float64 ($\beta{=}30$) & 1.0000 & 1.0000 & 0.9999 & 0.9983 & \cellcolor{gray!15}0.9449 & \cellcolor{gray!15}0.6565 \\
\bottomrule
\end{tabular}
\end{table}

\begin{figure}[t]
\centering
\includegraphics[width=0.75\linewidth]{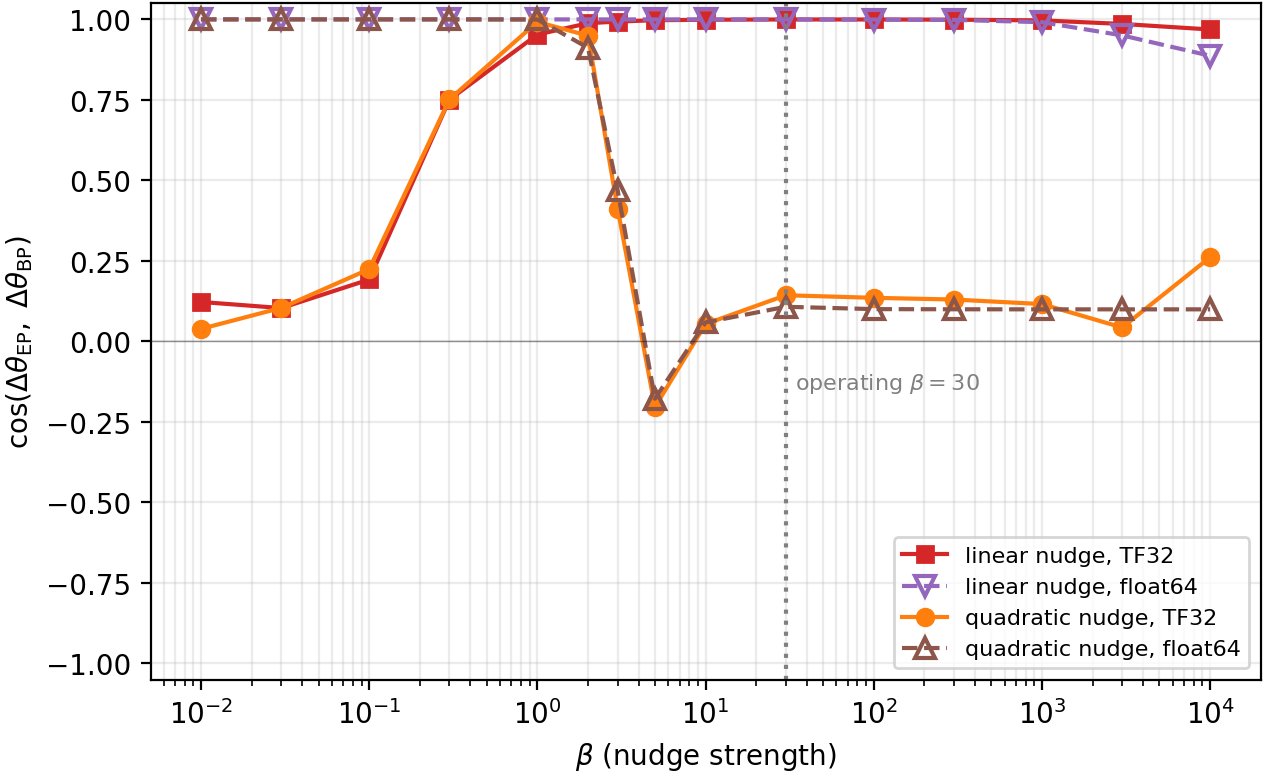}
\caption{$\cos(\Delta\theta_{\rm EP}, \Delta\theta_{\rm BP})$ against nudge
strength at $\gamma = 10^{-4}$, $512$ interpolants, $T_{\rm free} = 14$,
$T_{\rm nudge} = 6$, $K_h = 1$. The quadratic nudge's relaxation diverges above
$\beta = 1$; its values beyond that point are not true gradient estimates.}
\label{fig:train-beta}
\end{figure}

\paragraph{Nudge strength.} At the operating $\gamma = 10^{-4}$ the linear nudge holds $\cos \geq 0.99$ from $\beta = 3$ to $10^3$ and degrades only to
$0.969$ at $10^4$; below $\beta = 3$ it meets the TF32 floor, falling from
$0.95$ at $\beta = 1$ to about $0.1$ at $\beta = 0.01$, where float64 stays at
$1.0000$ throughout. The quadratic nudge's failure above $\beta = 1$
(Figure~\ref{fig:train-beta}) is a divergence of the nudged relaxation rather
than a loss of accuracy: the update norm leaves the backpropagation reference
of $0.323$ entirely, reaching $851$ at $\beta = 5$. It is a property of the
nudge form itself, not of the solver or the arithmetic: reducing the relaxation
step size at fixed nudge time makes it worse --- at $\beta = 3$ the norm grows
by a factor of nearly $400$ between $\mathrm{d}t = 1$ and $0.1$ --- and in
float64 the collapse occurs at the same $\beta$, the two arithmetics lying on
top of each other from $\beta=1$ upward. The linear nudge is unaffected by the step size, and shows no collapse in either arithmetic.

\paragraph{Solver and relaxation budgets.} The implicit solve's parameter
update is insensitive to the number of conjugate-gradient steps, giving
$\cos(\Delta\theta_{\rm IFT}, \Delta\theta_{\rm BP}) = 1.0000$ at every
$n_{\rm CG}$ from $1$ to $50$ and at every $\gamma$ tested, which is expected when the
relaxation Hessian is a perturbation of the identity
(App.~\ref{app:error-param}). The truncation therefore contributes nothing,
leaving the $O(\gamma)$ correspondence bias as the solve's only error. The parameter update also remains exact at the small $\gamma$
where Section~\ref{subsec:equivalence-at-inference} finds the inference
velocity arithmetic-limited, since a floor-level perturbation to
$v_{\rm PCN}$ barely rotates the direction of the parameter gradient it
produces. The two remaining EP budget axes, $T_{\rm nudge}$ and $K_h$, are likewise
flat from a single sweep (Table~\ref{tab:budget-axes}), for the deadbeat
reason of App.~\ref{app:error-param}, so $T_{\rm free}$ is the only budget
with a threshold, and the cost of an EP update is essentially the free-phase
equilibration alone.

\begin{table}[t]
\centering
\caption{Budget axes that are flat from a single sweep, at $\gamma = 10^{-4}$
on the CIFAR-10 replication checkpoint of Table~\ref{tab:ffn2pcn-fid}, $512$
flow-matching interpolants, otherwise the settings of our EP training run
($K_h = 1$, $T_{\rm free} = 14$, $T_{\rm nudge} = 6$, $\beta = 30$) with the
named axis varied. Left: conjugate-gradient steps for the implicit solve,
$\cos(\Delta\theta_{\rm IFT},\Delta\theta_{\rm BP})$; centre and right:
nudged-phase sweeps and hidden-equilibration steps for EP,
$\cos(\Delta\theta_{\rm EP},\Delta\theta_{\rm BP})$. Dashes mark values not in that axis's swept grid.}
\label{tab:budget-axes}
\begin{tabular}{r ccc}
\toprule
value & $n_{\rm CG}$ (IFT) & $T_{\rm nudge}$ (EP) & $K_h$ (EP) \\ \midrule
1 & 1.0000 & 0.9994 & 0.9998 \\
2 & 1.0000 & 0.9998 & 0.9998 \\
3 & 1.0000 & 0.9998 & 0.9998 \\
5 & 1.0000 & --- & 0.9998 \\
10 & 1.0000 & 0.9998 & --- \\
20 & 1.0000 & --- & --- \\
50 & 1.0000 & --- & --- \\
\bottomrule
\end{tabular}
\end{table}

\section{The hardware-plausible architecture}
\label{app:hardware-arch}

This appendix specifies the norm- and attention-free architecture of
Section~\ref{subsec:hardware} in enough detail to reimplement, together with
the training recipe it needed. It shows only that competitive generation is
possible on an architecture an analog substrate could run, not that this
particular architecture is the optimal one. We give our reasons for the
individual choices where we have them, but which of them are necessary, and at
what cost, we leave to future work.

\paragraph{What was removed and what replaced it.}
The architecture is built from the Energy Matching UNet but with three changes. Firstly, all
$51$ group-normalisation layers are replaced by the identity. Secondly, all $6$ attention
blocks are replaced by the identity, which must be done on the built module
tree rather than by configuration: the UNet's middle block hardcodes
ResBlock--Attention--ResBlock, so setting the attention resolutions alone
cannot produce an attention-free network. Thirdly, the ViT head is replaced by a flat one, which flattens the UNet's
$3\times32\times32$ output and applies $\mathrm{Linear}(3072, 512)$, SiLU,
$\mathrm{Linear}(512,1)$. We re-initialise the UNet's final convolution from $\mathcal{N}(0, 0.02)$
rather than leaving it at the flow-matching codebase's zero initialisation,
after an earlier variant of the head failed to train from zero. We did not
test whether the flat head trains without this.

Two further changes are made to the remaining convolutions and residual blocks. Firstly, weight standardisation~\citep{qiaoMicroBatchTrainingBatchChannel2020} replaces each
convolution's weights by
\begin{equation}
    \hat w_i = \frac{g_i\, w_i}{\sigma_i \sqrt{\mathrm{fan\text{-}in}} +
    \epsilon} , \qquad \sigma_i = \operatorname{std}(w_i) ,
\end{equation}
per output filter $i$, with $g_i$ a learned per-filter gain initialised to
$\sigma_i\sqrt{\mathrm{fan\text{-}in}}$ so that the reparameterisation is
exactly the identity at initialisation. No mean subtraction is applied. It is applied to every convolution except the zero-initialised ones --- therefore $42$ convolutions are
standardised and $23$ are left plain on the built model. Those $23$ are each
ResBlock's second convolution, zero-initialised by the flow-matching codebase
so that every block starts as the identity; standardising an all-zero filter
would leave both its weight and gain gradients at zero, so it could never leave
that initialisation. Secondly, residual scaling replaces each ResBlock's output by $\mathrm{skip}
+ \rho\,(f(x) - \mathrm{skip})$ with $\rho = 1/\sqrt{22} = 0.213$, one over
the square root of the number of ResBlocks~\citep{deBatchNormalizationBiases2020}. Neither requires computation at inference: $\hat w$ depends only on the weights, so
a crossbar is programmed with it once and never recomputes a normalisation at
runtime, and $\rho$ is a wiring gain.

The resolution-changing path needed no modification: downsampling is a
stride-$2$ convolution rather than pooling, upsampling is nearest-neighbour
interpolation followed by a stride-$1$ convolution, and both are weighted sums
like any other convolution.

\paragraph{Module census.}
The built model contains $42$ standardised convolutions, $23$ plain
convolutions, $22$ ResBlocks, $26$ linear layers, $69$ SiLU activations, and
zero group-normalisation, attention or layer-normalisation modules, against
$51$, $14$ and $16$ of those respectively in the architecture it replaces. It carries $78.4$M parameters against $50.0$M, from the flat head's
$3072\times512$ first linear layer and a base channel width of $192$ against
$128$, which widens every stage through the same multipliers.

\paragraph{How we arrived at this configuration.}
We reached it by trial rather than by systematic search. Attention and the ViT
head could each be removed without much cost, but removing normalisation made
training diverge, and weight standardisation was what made the fully stripped
network trainable. We did not establish which ingredients are necessary, at
what width, or under what schedule; that investigation is left to future work.

\paragraph{Training recipe.}
This arm is trained by backpropagation, with the settings of
Table~\ref{tab:hyperparams}. It differs from the other arms 
by using a cosine learning rate
schedule, annealing to zero at $145$k steps, then a constant learning rate of $6 \times 10^{-4}$ from $145$k to $175$k steps, after which the second phase is conducted as usual. We arrived at this change by trial after a constant learning rate of
$1.2\times10^{-3}$ destabilised partway through the first phase; we did not try
a lower constant rate from scratch, and the successful run spent most of its
life below $10^{-3}$, so the schedule rather than the annealing may be what
mattered.

\paragraph{Sampling time.}
The contrastive phase shifts the optimal sampling time in both architectures,
and further in this one. Sampled on a common $\tau_s$ ladder after that phase,
the full architecture (under all three training rules) is best at $\tau_s = 3.25$ and this one at $\tau_s = 5.0$,
and both are much worse at $\tau_s = 1$ than before the contrastive phase. Phase 2
optimises the potential for the Langevin regime at the expense of the noiseless
endpoint, and this architecture is affected more.

\section{Experimental implementation}
\label{app:experimental-implementation}

This appendix gives the settings needed to reproduce the runs of
Section~\ref{sec:cifar}, and the implementation choices that the PCN
formulation requires but a feedforward one does not. Code is available at
\url{https://github.com/alexgower/EnergyMatchingEqProp}.

\paragraph{Shared recipe.} Table~\ref{tab:hyperparams} gives the settings
shared by all arms, and the relaxation settings of the two PCN arms. The three
arms of Section~\ref{sec:cifar} share the architecture, optimiser, learning
rate, schedule, effective batch size, EMA decay, step budget and data pipeline;
the norm- and attention-free arm differs as App.~\ref{app:hardware-arch}
describes. What they do not share is the random realisation: there is no global
seed in the trainer, so each arm has its own initialisation, its own noise
samples for the optimal-transport coupling and the contrastive negatives, and
its own dropout masks. With one run per arm we have no estimate of run-to-run
variance, and the differences in Table~\ref{tab:cifar-fid} should be read with
that in mind. The implicit-solve arm holds $x$ fixed at $x_t$ and
differentiates the flow loss on $v_{\rm PCN}(x_t)$ with respect to $\theta$,
using the implicit function theorem for the dependence of $h^*$ on $\theta$,
so it uses no tether; the tethered objective $\mathcal{L}_{\rm PCN}$
approaches this fixed-$x$ one as $\gamma/\lambda \to 0$
(App.~\ref{app:training-details}).

\paragraph{Asynchronous relaxation.} Nodes are updated in even/odd sweeps
\citep{kerjanTrainingPredictiveCoding2026}, the second half-step seeing the
first's updated values, so one sweep costs two energy evaluations. On a chain
this is exactly Gauss--Seidel, since same-parity nodes share no energy term; on
the DAG of App.~\ref{app:dag-pcn} a skip may connect two same-parity nodes, so
the decoupling is only approximate. We adopt it because
\citet{kerjanTrainingPredictiveCoding2026} find synchronous updates
substantially worse at the same step size, and sometimes divergent; we did not
repeat that comparison here. The fixed point is the same either way.

\paragraph{Frozen dropout.} A PCN training step evaluates the network many
times: $T_{\rm free}$ sweeps during the free phase and $T_{\rm nudge}$ in each
nudged phase, each with $K_h$ inner sweeps, where a feedforward step evaluates
it once. Standard dropout would draw a fresh mask at every evaluation, so each
sweep would relax a different energy and there would be no single equilibrium
for the free phase to reach or for the finite difference to compare. We
therefore use one mask per sample, held fixed across every sweep and phase of
that sample's training step. It is drawn as standard dropout draws it, so only
the redrawing frequency differs, and for the feedforward arms --- a single
evaluation per step --- the two coincide.

\paragraph{FID protocol.} All FIDs are computed from 50k samples against the
full training set, with EMA weights, and Euler--Heun integration at
$\mathrm{d}t = 0.01$. The stopping time $\tau_s$ is chosen per model by
sampling over a ladder of values and taking the minimum: phase-1 checkpoints
are best at the noiseless endpoint $\tau_s = 1$, and after the contrastive
phase the optimum moves well past it (App.~\ref{app:hardware-arch}). The
$\tau_s = 3.25$ chosen as optimal for the backpropagation arm is also
near-optimal for both PCN arms, which a paired sweep of $\tau_s$ confirms. For
the ImageNet-32 checkpoints we use the authors' $\tau_s = 2.5$.

\paragraph{Measurement resolution and replication.} To gauge the measurement's
resolution we evaluated the backpropagation arm's phase-1 checkpoint at three
sampling seeds: the values span $0.064$. Changing how the 50k samples are
divided across devices moves a fixed-seed evaluation by a similar amount, and
unseeded draws of the same checkpoint span $0.15$. We therefore treat
differences below about $0.2$ as unresolved. Our replication reaches $3.54$
after phase 2 where the published model reports $3.34$; the authors' released
checkpoint scores $3.44$ under our protocol, so part of the gap is the sampler
and part the replication.

\paragraph{Relaxation step sizes.} The two blocks take different steps: the
errors are updated at $\mathrm{d}t = 1$, deadbeat up to $O(\gamma)$ for the
near-identity Hessian $I + \gamma M$ of the error parameterisation
(App.~\ref{app:error-param}), while the visible variables take
$\mathrm{d}t_x$, whose block carries the spring and the $O(\gamma)$ curvature
of $E_{\rm PCN}$ in $x$. Phase 1 uses $\mathrm{d}t_x = 1$; phase 2 uses
$\mathrm{d}t_x = 0.5$, paired with $\lambda = 2$ so that
$\lambda\,\mathrm{d}t_x = 1$ (below).

\paragraph{The contrastive phase.} The contrastive settings are those of
\citet{balcerakEnergyMatchingUnifying2025}: $200$ Langevin steps at
$\mathrm{d}t = 0.01$ and $\epsilon_{\max} = 0.01$, half the chains initialised
from data and half from noise on the same noise schedule, entering the loss with weight
$\lambda_{\rm CD} = 10^{-3}$ as a trimmed mean over the negatives with the top
$10\%$ dropped and the term clamped at $0.02$.

For the PCN arms, negatives are drawn by a
Langevin chain on a feedforward copy of the network carrying the PCN's current
weights, rather than by relaxation, which our compute budget did not extend to;
no parameter gradient passes through that copy, and
Section~\ref{subsec:equivalence-at-inference} shows the two velocity fields
agree to a cosine similarity of $1$ to nine decimal places. The EP arm's contrastive phase starts from its $100$k phase-1 checkpoint; its
$145$k checkpoint did not survive the first contrastive updates under any
relaxation setting we tried. The failure is localised to the ViT head: its
attention saturates between $130$k and $140$k, and restoring the affected
layers from $130$k lets the $145$k checkpoint train. From $100$k, the phase-1 relaxation settings did
not carry over either: the hidden relaxation stopped converging on some batches
after about $60$ contrastive steps. The contrastive term raises the flow
residual $|g|$, which enlarges the nudge displacement
$\beta\gamma|g|/\lambda^2$ of \eqref{eq:lin-displacement} and tightens
\eqref{eq:lin-constraint}; the phase-2 settings reduce its prefactor
$\beta\gamma/\lambda^2$ by four to compensate. $\gamma$ falls by $33\times$
and $\beta$ rises by the same factor, keeping $\beta\gamma$ fixed, and the
tether stiffens to $\lambda = 2$, paired with $\mathrm{d}t_x = 0.5$ so that
$\lambda\,\mathrm{d}t_x = 1$ remains the deadbeat choice of
Section~\ref{subsec:training}. What matters is the reduced prefactor, not how
it is achieved. In short contrastive runs from the same checkpoint, every
setting at the phase-1 prefactor lost relaxation convergence within about
$100$ steps, whatever its $\lambda$, including the stiffer tether with $\beta$
raised to restore the phase-1 prefactor. Every setting with the prefactor reduced by a factor of four
trained cleanly, among them the simplest choice, $\beta$ reduced by a factor of four with
$\gamma$, $\lambda$ and $\mathrm{d}t_x$ at their phase-1 values, which ran
$200$ contrastive steps with the relaxation converged in every phase.

\begin{table}[t]
\centering
\caption{Phase-1 FID against training step, from 50k samples with EMA weights
at $\tau_s = 1.0$. All three arms improve substantially to $100$k and are
slightly worse by $145$k under the constant learning rate, and the EP arm's
$100$k checkpoint is better than any arm's $145$k, so its contrastive phase
does not start from a weaker model. Repeated evaluations of one checkpoint
span up to $0.15$, and we treat differences below about $0.2$ as unresolved.}
\label{tab:phase1-fid}
\begin{tabular}{lrrr}
\toprule
 & \multicolumn{3}{c}{Phase-1 FID $\downarrow$} \\
\cmidrule(lr){2-4}
Gradient & $50$k steps & $100$k steps & $145$k steps \\
\midrule
Backpropagation & 11.22 & 6.13 & 6.42 \\
Implicit solve & 12.35 & 6.45 & 6.70 \\
EP & 12.60 & 6.23 & 6.87 \\
\bottomrule
\end{tabular}
\end{table}

\begin{table}[t]
\centering
\caption{Training settings. The upper block is shared by all arms; the lower
gives the relaxation settings of the two PCN arms, with EP's two phases
separately (sweep budgets defined in App.~\ref{app:training-details}).}
\label{tab:hyperparams}
\begin{tabular}{llll}
\toprule
\multicolumn{4}{l}{\itshape shared} \\
Optimiser & \multicolumn{3}{l}{Adam, $\beta = (0.9, 0.95)$, gradient clipping at norm $1$} \\
Learning rate & \multicolumn{3}{l}{$1.2\times10^{-3}$, linear warmup over $10$k steps} \\
Effective batch & \multicolumn{3}{l}{$1024$ (phase 1), $512$ (phase 2)} \\
EMA decay & \multicolumn{3}{l}{$0.9999$ (phase 1), $0.99$ (phase 2)} \\
Steps & \multicolumn{3}{l}{$145$k (phase 1; EP's contrastive phase starts from
    its $100$k checkpoint) $+\,2$k (phase 2)} \\
Flow objective & \multicolumn{3}{l}{$u = x_1 - x_0$, exact OT per batch, $\sigma = 0$} \\
Contrastive phase & \multicolumn{3}{l}{$200$ Langevin steps, $\mathrm{d}t = 0.01$,
    $\epsilon_{\max} = 0.01$, $\lambda_{\rm CD} = 10^{-3}$} \\
\midrule
\multicolumn{4}{l}{\itshape relaxation settings} \\
 & implicit solve  & EP (phase 1) & EP (phase 2) \\
$\gamma$            & $10^{-4}$ & $10^{-4}$ & $3\times10^{-6}$ \\
$\lambda$           & ---       & $1$       & $2$ \\
$\beta$             & ---       & $30$      & $10^{3}$ \\
$K_h$               & $2$       & $1$       & $1$ \\
$T_{\rm free}$      & ---      & $14$      & $14$ \\
$T_{\rm nudge}$     & ---       & $6$       & $6$ \\
nudge               & ---       & linear    & linear \\
$\mathrm{d}t_x$     & ---       & $1$       & $0.5$ \\
$\mathrm{d}t$ (errors)    & $1$       & $1$       & $1$ \\
$n_{\rm CG}$        & $10$      & ---       & --- \\
\bottomrule
\end{tabular}
\end{table}

\section{Scope of the claims}\label{app:scope}

This appendix states, in one place, what the paper does and does not claim,
on four points where a reader might reasonably read more into the results
than the evidence supports.

\paragraph{The solver is not the substrate.} Every equilibrium in this paper
was found by gradient descent in error coordinates (App.~\ref{app:error-param}),
and a single deadbeat sweep of that solver is algebraically a backward pass
\citep[Thm.~C.9]{goemaereEPCOvercomingExponential2025}. That is a choice of
simulation method: the claim of Section~\ref{subsec:correspondence} is a
property of the equilibrium, not of the route to it. The local state dynamics
\eqref{eq:dag-dynamics} share that equilibrium exactly and converge towards
it (App.~\ref{app:error-param}), but too slowly under explicit discretisation
to run at this scale. What the paper establishes is that a state whose local
errors carry $-\nabla_x V$ exists, that the parameter updates EP reads from
it match backpropagation's, and that training on those updates works at
CIFAR-10 scale. What it does not establish is that a physical system settles
there in useful time; that is the open question a substrate would answer.

\paragraph{What is new.} The paper builds on known ingredients: that a PCN's
equilibrium errors implement the backward pass
\citep{whittingtonApproximationErrorBackpropagation2017,
millidgePredictiveCodingApproximates2022}, the spring-clamped free phase of
GradEP \citep{gowerFlowEqPropTrainingFlow2026}, and the linearised nudge of
Dual Propagation \citep{hoierDualPropagationAccelerating2023}. Its
contributions are the injection-term construction that turns a
scalar-output network into one whose visible layer carries $-\nabla_x V$
(Proposition~\ref{prop:correspondence}); the joint $\gamma$--$\beta$ window
and its derivation (App.~\ref{app:linearised-nudge},
Table~\ref{tab:gamma-beta}); and the demonstration that these compose into
competitive generation on CIFAR-10, where prior backpropagation-free
generative models remain at MNIST scale.

\paragraph{Where a backward pass remains in training.} The EP arm's parameter
updates in both phases come from measurements at equilibrium. Its phase-2
negatives, however, were sampled by a feedforward twin of the network using
autograd velocities, since relaxing $200$ Langevin steps per negative was
beyond our budget. No parameter gradient passes through the twin, and the two
velocity fields agree to a cosine similarity of $1$ to nine decimal places
(Section~\ref{subsec:equivalence-at-inference}), so we expect replacing it with relaxation changes the
negatives by an amount below the sampler's own noise, though we did not test this. ``Without
backpropagation'' in Section~\ref{sec:cifar} refers to how the weights were
updated, not to every computation the training loop performed.

\paragraph{The hardware-plausible architecture is an existence proof.} The
norm- and attention-free network of Section~\ref{subsec:hardware} is wider
than the reference, used a different schedule, and was trained and sampled
feedforward only. Its FID of $3.51$ shows that neither component is needed
for competitive generation; it does not show that they can be removed at
matched cost, nor that the resulting network trains under EP. Both are left
to future work, as is the run-to-run variance of every arm in
Table~\ref{tab:cifar-fid}, each of which is a single run.

\section{Generated samples}\label{app:samples}

\begin{figure}[t]
\centering
\includegraphics[width=\linewidth]{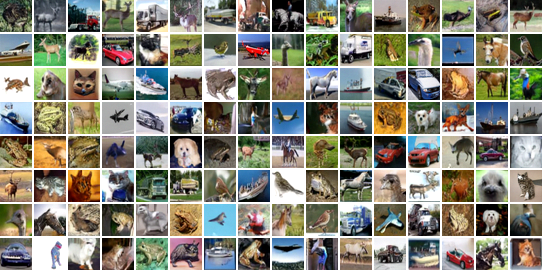}
\caption{CIFAR-10 samples from the backpropagation arm, FID $3.54$ at
$\tau_s = 3.25$.}
\label{fig:samples-bp}
\end{figure}

\begin{figure}[t]
\centering
\includegraphics[width=\linewidth]{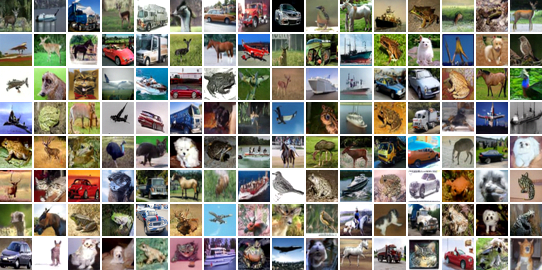}
\caption{CIFAR-10 samples from the implicit-solve arm, FID $3.64$ at
$\tau_s = 3.25$. Parameter gradients come from implicit differentiation of the
PCN's equilibrium rather than from backpropagation through a forward pass.}
\label{fig:samples-ift}
\end{figure}

\begin{figure}[t]
\centering
\includegraphics[width=\linewidth]{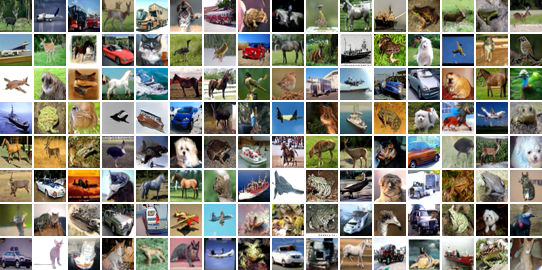}
\caption{CIFAR-10 samples from the EP arm, FID $3.64$ at $\tau_s = 3.25$ ---
a model trained without backpropagation in either phase.}
\label{fig:samples-ep}
\end{figure}

\begin{figure}[t]
\centering
\includegraphics[width=\linewidth]{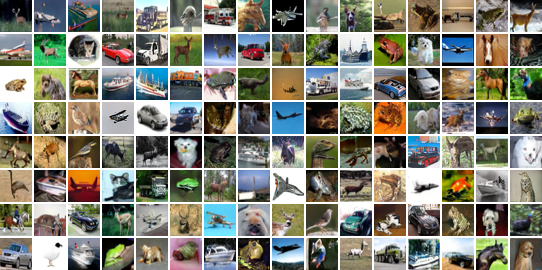}
\caption{CIFAR-10 samples from the norm- and attention-free architecture of
Section~\ref{subsec:hardware}, FID $3.51$ at $\tau_s = 5.0$. Every operation in
this network is a fixed weighted sum or a pointwise nonlinearity.}
\label{fig:samples-ws192}
\end{figure}

Figures~\ref{fig:samples-bp}--\ref{fig:samples-ws192} show $128$ samples from
each of the four trained models of Section~\ref{sec:cifar}, drawn from each
model's phase-2 checkpoint under its own native inference --- feedforward
autograd for the two backpropagation-trained models, PCN relaxation for the
other two. All four share a sampling seed, so the panel in a given position
starts from the same $x_0$ in every figure; the sampling time is each model's
own optimum and is given per figure.

\end{document}